%% file: 0_main.tex
\documentclass[10pt,twocolumn,letterpaper]{article}

\usepackage{cvpr}              

\usepackage{graphicx}
\usepackage{amsmath}
\usepackage{amsfonts}
\usepackage{amssymb}
\usepackage{booktabs}
\usepackage{algpseudocode}
\usepackage{algorithm}
\usepackage[table,x11names,dvipsnames]{xcolor}
\usepackage{pifont}
\definecolor{paleRed}{RGB}{255,220,220}

\newcommand{\cmark}{\ding{51}}%
\newcommand{\xmark}{\ding{55}}%

\newcommand*{\modelname}{Koala\xspace}

\newcommand*{\modelabb}{Koala\xspace}

\usepackage[pagebackref,breaklinks,colorlinks]{hyperref}

\usepackage[capitalize]{cleveref}
\crefname{section}{Sec.}{Secs.}
\Crefname{section}{Section}{Sections}
\Crefname{table}{Table}{Tables}
\crefname{table}{Tab.}{Tabs.}

\def\cvprPaperID{5177} 
\def\confName{CVPR}
\def\confYear{2024}

\newcommand\blfootnote[1]{%
  \begingroup
  \renewcommand\thefootnote{}\footnote{#1}%
  \addtocounter{footnote}{-1}%
  \endgroup
}

\begin{document}

\title{\raisebox{-0.1\height}{\includegraphics[height=1em]{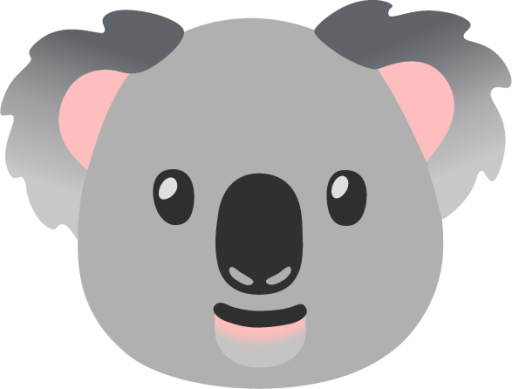}} Koala: Key frame-conditioned long video-LLM}
%

\author{Reuben Tan$^{1}$ \ \ \ \ Ximeng Sun$^{1}$ \ \ \ \ Ping Hu$^{1*}$ 
\ \ \ \ Jui-hsien Wang$^{2}$ \ \ \ \  Hanieh Deilamsalehy$^{2}$ \ \ \ \  \\
Bryan A. Plummer$^{1}$ \ \ \ \ Bryan Russell$^{2}$ \ \ \ \ Kate Saenko$^{1}$ \\
$^{1}$Boston University, $^{2}$Adobe Research \\
{\tt \small \{rxtan, sunxm, pinghu, bplum, saenko\}@bu.edu}, {\tt \small \{juiwang, deilamsa, brussell\}@adobe.com} \\
\url{https://cs-people.bu.edu/rxtan/projects/Koala} \\
}

\twocolumn[{%
\renewcommand\twocolumn[1][]{#1}%
\maketitle%
\vspace{-0.3in}%
\begin{center}
\includegraphics[width=\linewidth]{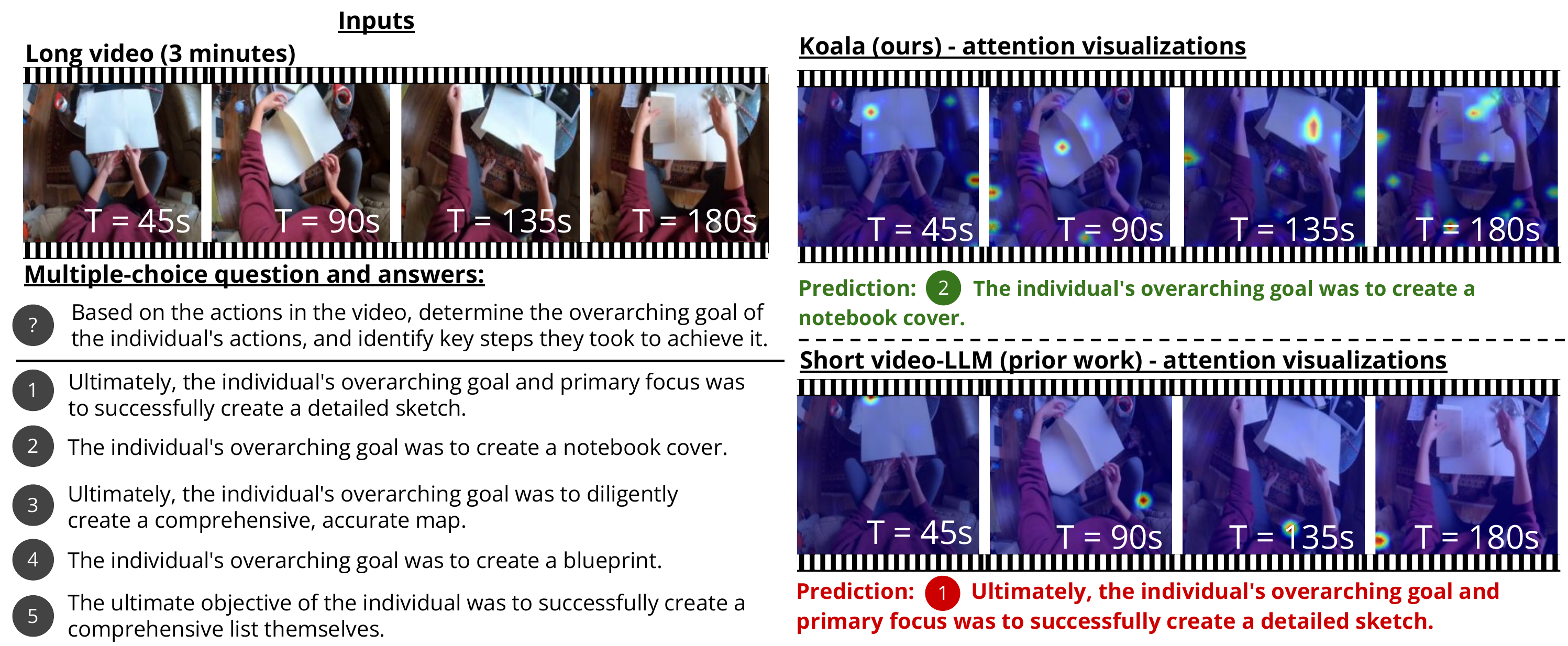}
        \vspace{-0.2in}
   \captionof{figure}{Given a video-Large Language Model that was pretrained on millions of short \emph{seconds}-long video clips, we propose a lightweight approach (\modelabb) to extend its short-term video tokenizer function for understanding and  answering questions about \emph{minutes}-long videos. We are the first to use sparsely sampled key frames to condition the LLM. As shown, our \modelabb approach is more effective at focusing on relevant regions in the input frames than the short vLLMs, allowing it to make more informed predictions based on a more holistic understanding of the video. These regions help facilitate our model in predicting the correct answer to the question (highlighted in green).
   } 
\label{fig:motivational_figure}
\end{center}
}]

\maketitle

\begin{abstract}
Long video question answering is a challenging task that involves recognizing short-term activities and reasoning about their fine-grained relationships. State-of-the-art video Large Language Models (vLLMs) hold promise as a viable solution due to their demonstrated emergent capabilities on new tasks. However, despite being trained on millions of short \emph{seconds}-long videos, vLLMs are unable to understand \emph{minutes}-long videos and accurately answer questions about them. To address this limitation, we propose a lightweight and self-supervised approach, Key frame-conditioned long video-LLM (Koala), that introduces learnable spatiotemporal queries to adapt pretrained vLLMs for generalizing to longer videos. Our approach introduces two new tokenizers that condition on visual tokens computed from sparse video key frames for understanding short and long video
moments. We train our proposed approach on HowTo100M and  demonstrate its effectiveness on zero-shot long video understanding benchmarks, where it outperforms state-of-the-art large models by 3 - 6\% in absolute accuracy across all tasks. Surprisingly, we also empirically show that our approach not only helps a pretrained vLLM to understand long videos but also improves its accuracy on short-term action recognition. 

\end{abstract}
\vspace{-15pt}
\blfootnote{*Currently at the University of Electronic Science and Technology of China}
\input{1_introduction.tex}
\input{2_related.tex}
\input{3_approach.tex}
\input{4_experiments.tex}
\input{5_conclusion.tex}

{\small
\bibliographystyle{ieee_fullname}
\bibliography{egbib}
}

\clearpage
\appendix
\input{6_supplemental.tex}


\end{document}

%% file: 1_introduction.tex
\section{Introduction}
Answering questions about minutes-long videos  is an inherently challenging task that involves recognizing multiple  actions and how they fit together to form the overall activity. To recognize that the person is making a notebook cover instead of a sketch in Figure~\ref{fig:motivational_figure}, a model must spot key actions (taping, measuring) and objects (paper), and understand how they are related to each other.
Instruction-tuned multimodal-Large Language Models (mLLMs) \cite{liu2023visual,zhu2023minigpt,li2023blip,instructblip,zhang2023transfer} and their video variants (vLLMs) \cite{luo2023valley,li2023videochat,zhang2023video,maaz2023video} offer a promising avenue for understanding long videos, as demonstrated by their emergent capabilities in downstream multimodal tasks including perception \cite{tsimpoukelli2021multimodal} and commonsense reasoning \cite{wei2022emergent,instructblip}. By learning to tokenize a small number of key frames from \emph{seconds}-long videos into visual tokens that are mapped to the same latent space as language word tokens, vLLMs are able to leverage the knowledge encapsulated in their LLM to describe visual concepts, such as actions, in short videos. 

However, existing vLLMs trained on millions of short videos still struggle with \emph{minutes}-long videos that contain significantly more frames \cite{li2023seed}. A naive solution is to extract the same number of key frames at a coarse rate, but this leads to a significant loss of fine-grained spatiotemporal information.  Thus, this approach results in poor performance on complex and long-term temporal understanding tasks in benchmarks including EgoSchema \cite{mangalam2023egoschema} and Seed-Bench \cite{li2023seed}.
Another possibility for extending these pretrained vLLMs to long videos is to pass multiple segments of key frames into their learned tokenizer function. However, this extension may negatively affect the ability of the vLLMs to understand long videos holistically since their tokenizer function only aggregates spatiotemporal context \emph{within} segments rather than \emph{between} them.


In light of these limitations, we propose our Key frame-conditioned long video-LLM (\modelname), a novel and self-supervised approach that introduces spatiotemporal queries to adapt the \emph{frozen} video tokenizer in pretrained vLLMs to aggregate spatiotemporal context over longer temporal horizons. 
Our main hypothesis is that the video tokenizer function in vLLMs, having learned to aggregate spatiotemporal context for a fixed number of frames, can generalize to understanding longer videos using the same number of input frames. More specifically, we first encode the global context of a long video by extracting the same number of input frames at a very coarse sampling rate, referred to as \textit{key frames}. To mitigate the loss of fine-grained spatiotemporal information, we then extract a sequence of video segments at a higher sampling rate to complement the global context with local spatiotemporal information.  

The key insight underlying \modelabb is that the global video context can be utilized to model individual video segments and the contextual relations \emph{between} multiple video segments, which plays a crucial role in understanding long videos. 
To this end, we further introduce our Conditioned Segment (CS) and Conditioned Video (CV) tokenizer functions. 
Intuitively, the former function leverages learnable segment queries that use the global context of the video to identify and aggregate frame-level concepts within each segment; such concepts are important to both short-term context of the segment and the global context of the entire video. 
The latter function further introduces temporal concept queries to reason about the contextual relationships between segments to generate an enriched sequence of visual tokens as inputs into the subsequent LLM.

While the idea of using frames extracted at different sampling rates bears similarities to existing approaches \cite{li2022weakly,kahatapitiya2021coarse,sun2022coarse} including slowfast network \cite{feichtenhofer2019slowfast}, these aforementioned approaches focus on modeling static and motion contexts in \emph{short} videos, especially in a closed-world setting. In contrast, we focus on a task-agnostic approach for computing enriched visual tokens that are well-aligned with the base LLMs. More significantly, reasoning about global and short-term semantics of videos in vLLMs makes our setting different and challenging. 
By facilitating long video understanding with LLMs, our \modelabb approach helps to address the inherent problem of summarizing and understanding high-level temporal context which is prevalent in downstream open-world applications including video recommendations \cite{gorti2022x, liu2022ts2, hu2022lightweight}, embodied AI \cite{deitke2022️, li2023behavior, khandelwal2022simple,gadre2022continuous} and robotics \cite{kumar2023graph, qin2022dexmv}.


We demonstrate the effectiveness of our proposed \modelabb approach through extensive evaluations on multiple zero-shot long and short-term temporal understanding tasks on the EgoSchema \cite{mangalam2023egoschema} and the Seed-Bench \cite{li2023seed} benchmarks. We show that our proposed light-weight finetuning approach is able to incorporate long-term temporal understanding capabilities into pretrained vLLMs despite training on noisy and uncurated video and text data from the Howto100M dataset \cite{miech2019howto100m}, and outperforms state-of-the-art mLLMs by a significant margin of 3 - 6\% across all tasks. Furthermore, we show that our CS and CV tokenizer functions also help the base vLLM to improve its performance on short-term action recognition. We provide a comprehensive ablation of our approach to analyze the effectiveness of the spatiotemporal queries introduced in the proposed tokenizer functions in \modelabb. We are the first work to explore extending the video tokenizer function of pretrained short-term vLLMs to long-term video understanding.

%% file: 2_related.tex
\section{Related work}
\begin{figure}[t]
\vspace{-5mm}
\begin{center}
\includegraphics[width=1.0\linewidth]{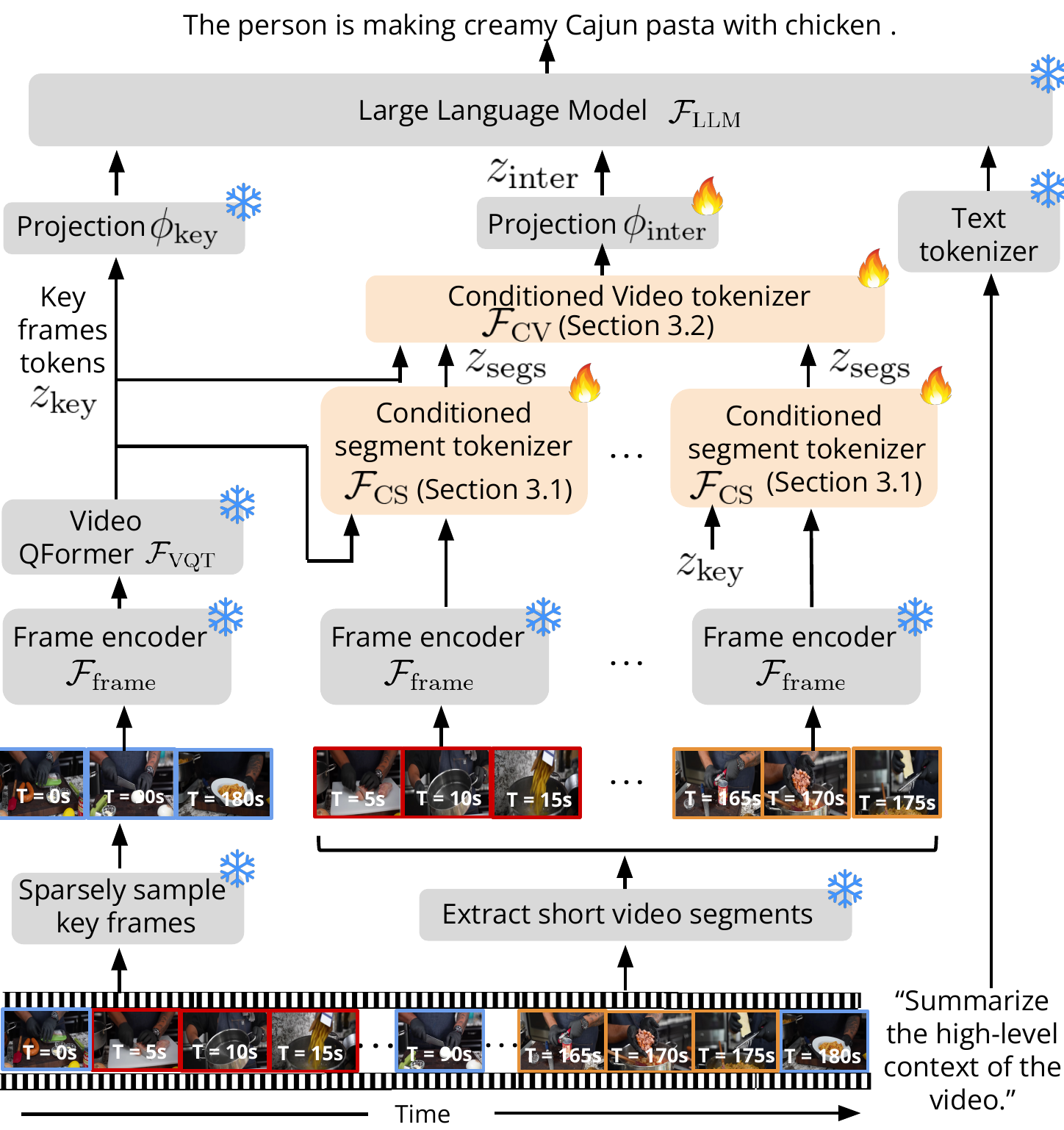}
\end{center}
\vspace{-4mm}
   \caption{ \textbf{Overview of our full \modelabb approach.} For a given video, we extract a set of coarsely-sampled key frames and non-overlapping frame segments with a much higher sampling rate. We use the \textcolor{blue}{key frames} to provide high-level global context of the video to compute a final sequence of soft visual tokens that encode both global context as well as fine-grained spatiotemporal information via the Conditioned Segment (CS) and Conditioned Video (CV) tokenizer functions.} 
\label{fig:model_overview}
\vspace{-3mm}
\end{figure}

\noindent\textbf{Video understanding.} The field of video understanding encompasses core research problems including action recognition \cite{feichtenhofer2021large,wu2022memvit}, action prediction \cite{girdhar2021anticipative} and temporal action localization \cite{li2022weakly}. Older prior work addressing these problems are often task-specific and either rely on hand-crafted features \cite{klaser2008spatio,wang2013dense, yuan2009discriminative} or video encoders that are carefully designed to exploit temporal information from RGB frames and optical flow information \cite{carreira2017quo,tran2018closer,feichtenhofer2019slowfast,feichtenhofer2020x3d}. Moreover, understanding action sequences has often been constrained to short video clips. vLLMs are also similar to more recent fully attentional video encoders \cite{arnab2021vivit, bertasius2021space, fan2021multiscale,neimark2021video} that leverage self-attention between spatiotemporal regions to compute more effective video representations. 

Additionally, there are also existing works which aim to address the task of understanding long videos \cite{wu2021towards,wang2023selective}. While these aforementioned approaches are similar in spirit to our proposed approach, they are focused on recognizing actions instead of generating language as in our case.

\noindent\textbf{Instruction-tuning and multimodal foundation models.} Recently, instruction-tuned multimodal-LLMs \cite{instructblip,li2023videochat,luo2023valley, zhang2023transfer, zhang2023video} have demonstrated surprising emergent capabilities on unseen tasks. We make the distinction between two main types of multimodal LLMs - image-based \cite{alayrac2022flamingo,zhu2023minigpt,liu2023visual, zhang2023transfer,su2023pandagpt,wang2023visionllm} and video-based \cite{zhang2023video, luo2023valley,maaz2023video, li2023videochat, ye2023mplug}. In general, mLLMs learn an adaptor between the frozen visual encoders and the LLM that generates a sequence of soft visual tokens. The base LLMs are often kept frozen or lightly finetuned with the LORA framework \cite{hu2021lora} to leverage their vast amount of knowledge gleaned from large-scale pretraining \cite{touvron2023llama, vicuna2023,chung2022scaling,zhang2022opt}. While our proposed \modelabb model is also built upon a base vLLM, a key difference between prior mLLMs and ours lies in the way temporal information is aggregated in the video domain. Prior vLLMs \cite{maaz2023video, zhang2023video, luo2023valley} are often pretrained on large-scale and publicly available video and text datasets, as well as a highly curated instructional video dataset that has been annotated with temporal and spatial relations by Chat-GPT \cite{ChatGPT}. However, despite tuning on this dataset, state-of-the-art video-LLMs are still limited at understanding temporal relationships. Furthermore, while there are existing multimodal approaches \cite{sun2022long,gao2023mist} that have also been introduced to address the task of long video question answering, they differ from ours in different aspects. \cite{gao2023mist} conditions
the computation of visual attention on the question but ours uses global visual context. \cite{sun2022long} relies on fine-grained paragraph annotations while ours only relies on coarse and noisy
goal labels.

\noindent\textbf{Comparisons to existing prompting approaches.} \modelabb is similar in spirit to existing approaches that use learnable queries for foundational image-text models \cite{radford2021learning} for short-term action recognition \cite{ni2022expanding, ju2022prompting, wasim2023vita}. However, their purpose is introducing temporal prompts to transform the learned spatial aggregation function to reason about the temporal relations between a small number of frames. In contrast, we use spatiotemporal prompts to extend the learned short-term temporal aggregation function for long-term understanding of videos at least 10 times longer. Furthermore, our proposed approach provides an efficient mechanism for aggregating long-term temporal context over multiple segments.

%% file: 3_approach.tex
\section{Koala}
\begin{figure*}[h]
\vspace{-5mm}
\begin{center}
\includegraphics[width=0.95 \linewidth]{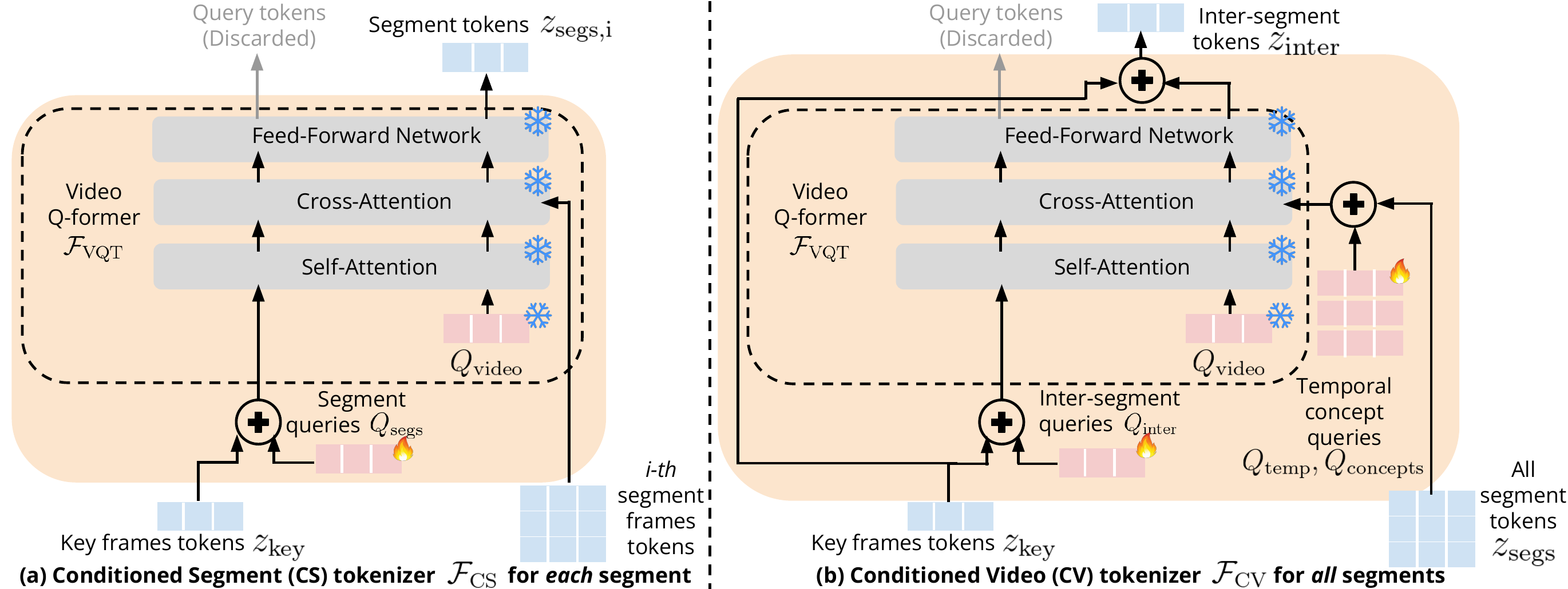}
\end{center}
\vspace{-6mm}
   \caption{\textbf{CS and CV tokenizer functions.} (a) Our CS tokenizer introduces learnable segment queries and fuses the global semantics of a video with fine-grained frame concept representations within each segment to compute segment tokens. (b) In the CV module, we introduce learnable inter-segment queries as well as temporal concept queries to model the contextual relations between segments.} 
\label{fig:global_frame_segment_attn}
\vspace{-3mm}
\end{figure*}

We propose \modelname, a lightweight finetuning approach that takes a frozen vLLM, which is pretrained on short video clips, and adapts it to longer temporal settings. The key components of \modelname are visual tokenizers that condition on representations of a sparse set of video key frames to adaptively select and aggregate information at the \emph{segment} and \emph{video} levels. We assume the vLLMs \cite{zhang2023video, li2023videochat} are trained to generate a textual response that is conditioned on an input text query and a short (seconds-long) video. The input text query is encoded to a set of text tokens $z_\text{text}$. To encode the video $V$, the pretrained vLLM samples a fixed number of key frames $V_\text{key}\subset V$, and then applies a key frames tokenizer function $\mathcal{F}_\text{key}$.  $\mathcal{F}_\text{key}$ aggregates the spatiotemporal context over the visual features within the set of key frames and returns a set of key frames tokens $z_\text{key}$.    

Let $z_\text{key} = \mathcal{F}_\text{key}(V_{\text{key}}) = \mathcal{F}_{\text{VQT}}(\mathcal{F}_{\text{frame}}(V_{\text{key}}); Q_\text{video})$, where $\mathcal{F}_\text{VQT}$ and $\mathcal{F}_\text{frame}$ denote pretrained video QFormer \cite{zhang2023video,li2023blip} and frame encoding functions, respectively. Similar to the Perceiver model \cite{jaegle2021perceiver}, $\mathcal{F}_\text{VQT}$ is partly parameterized by a set of frozen video queries $Q_{\text{video}}$ (\cf, Figure~\ref{fig:global_frame_segment_attn}) for aggregating the spatiotemporal information within $V_\text{key}$. In this work, we term the information encoded by $z_\text{key}$ as the global context of the video.
Given the sets of text and key frames tokens $z_\text{text}$ and $z_\text{key}$, respectively, the LLM function $\mathcal{F}_{\text{LLM}}$ computes the output textual response $r$ as: 
\small
\begin{equation} \label{eq:llm_func}
    r = \mathcal{F}_{\text{LLM}}(\text{concat}\{z_\text{text}, \phi_\text{key}(z_\text{key}) \}),
\end{equation}
\normalsize
where concat\{\} is the concatenation operation and $\phi_\text{key}$ is an affine transformation that projects the visual tokens to the LLM token space.

While the key frames tokenizer $\mathcal{F}_{\text{key}}$ encodes the global context of a long video by reasoning about the high-level relationships between key frames, the coarse sampling rate results in a loss of fine-grained spatiotemporal information that is crucial for understanding long videos effectively. To address this limitation, we propose to enrich the key frames tokens with the spatiotemporal information of \emph{local} video segments, illustrated in Figure~\ref{fig:model_overview}. Specifically, we compute a set of contextualized inter-segment tokens $z_\text{inter}$ from $N$ non-overlapping video segments $S = \{S_1, \cdots, S_{N} \}$, where each segment $S_i\subset V$ is sampled at a higher frame rate. We modify Eq~(\ref{eq:llm_func}) to include the inter-segment tokens $z_\text{inter}$ and the learnable affine transformation $\phi_\text{inter}$:
\small
\begin{equation}
    r = \mathcal{F}_{\text{LLM}}(\text{concat}\{z_\text{text}, \phi_\text{key}(z_\text{key}), \phi_\text{inter}(z_\text{inter}) \}).
\end{equation} 
\normalsize
To compute $z_\text{inter}$, we introduce our Conditioned Segment (CS) and Conditioned Video (CV) tokenizer functions, which repurpose the \emph{frozen} $\mathcal{F}_\text{VQT}$ function to select local spatiotemporal information that are most relevant to the conditioned global context at the \emph{segment} and \emph{video} levels.

\textbf{At the segment level}, our CS tokenizer $\mathcal{F}_{\text{CS}}$ (Section~\ref{sec:hier_spatio}) uses learnable queries that are conditioned on the encoded global context of $z_{\text{key}}$ to identify visual concepts in frames. We seek visual concepts that are not only relevant to the local context within each segment, but also to the global context of the entire video. This context is needed because $\mathcal{F}_{\text{VQT}}$ only aggregates the contextual relationships \emph{within} segments of frames and not \emph{between} them. 
\textbf{At the video level}, our CV tokenizer function $\mathcal{F}_{\text{CV}}$ (Section~\ref{sec:segment-attn}) leverages $\mathcal{F}_\text{VQT}$ to reason about the contextual relationships of spatiotemporal concepts across different segments conditioned on the global context of $z_\text{key}$. 
Taken together, the final sequence of contextual inter-segment tokens $z_\text{inter}$ is the output of the composition of these tokenizers:
\small
\begin{equation}
    z_\text{inter} = \mathcal{F}_\text{CV}(\{\mathcal{F}_\text{CS}(S_i \ | \ z_\text{key})\}_{i=1}^N \ | \ z_\text{key})
\end{equation} 
\normalsize
Note that the attention mechanism encapsulated by the CS and CV tokenizers facilitates the dissemination of global video context to more fine-grained visual concepts. 
Finally, we describe our learning objective in Section~\ref{sec:learning-obj}.

\subsection{Conditioned Segment Tokenizer} \label{sec:hier_spatio}
We illustrate our Conditioned Segment (CS) tokenizer in Figure~\ref{fig:global_frame_segment_attn}\textcolor{red}{a}. This tokenizer repurposes the key frames tokenizer $\mathcal{F}_\text{key}$ to select important frame-level information that is pertinent to both the local context of each segment and the global context of the key frames tokens $z_\text{key}$. 
As we will demonstrate empirically in Section~\ref{sec:ablations}, naively increasing the number of key frames as input into $\mathcal{F}_{key}$ during finetuning does not help the vLLM to generalize to longer videos, even when accounting for the quadratic complexity of the attention operation. 

For video segment $S_i \in S$, we repurpose the key frames tokenizer $\mathcal{F}_\text{key}$ via two simple modifications to the video QFormer $\mathcal{F}_{\text{VQT}}$. First, we concatenate the key frame tokens $z_\text{key}$ with the video QFormer's pretrained video queries $Q_\text{video}$. This modification allows the video QFormer to condition on the key frame tokens when aggregating the input video segment features $\mathcal{F}_\text{frame}(S_i)$ via cross-attention with $Q_\text{video}$ and $z_\text{key}$. Second, to ensure that the key frame tokens $z_\text{key}$ are compatible with the video QFormer, we adapt them via addition with a set of learnable queries $Q_\text{segs}$. 
For video segment $S_i$, we define the CS tokenizer $\mathcal{F}_{\text{CS}}$ as:
\small
\begin{equation}
    \mathcal{F}_\text{CS}(S_i \ | \ z_\text{key}) = \mathcal{F}_\text{VQT}(\mathcal{F}_\text{frame}(S_i); \text{concat}\{Q_\text{video}, z_\text{key} + Q_\text{segs}\}).
\end{equation}
\normalsize
Note that this CS tokenizer outputs tokens for $Q_\text{video}$ and $z_\text{key}$. We empirically find that it is beneficial to discard the output tokens for $z_\text{key}$. 

\subsection{Conditioned Video Tokenizer} \label{sec:segment-attn}
While our CS tokenizer helps to augment the local context of segment tokens with the global context of the entire video, the resulting tokens for each segment still lack contextual information from other segments. As such, we further propose our Conditioned Video (CV) tokenizer $\mathcal{F}_\text{CV}$ to reason about important spatiotemporal relationships \emph{across} segments  (Figure~\ref{fig:global_frame_segment_attn}\textcolor{red}{b}).

\noindent\textbf{Modeling spatiotemporal context across segments.} We model how the local segments are related to each other conditioned on the global context of the entire video. This objective involves a granular understanding of how specific concepts such as entities and action sequences are interconnected throughout the entire video. Let $z_{\text{segs}, i} = \mathcal{F}_\text{CS}(S_i \ | \ z_\text{key})$ be the set of conditioned tokens for segment $S_i$. To ensure that these tokens are compatible with the video QFormer $\mathcal{F}_\text{VQT}$, we introduce a set of learnable temporal queries $Q_{\text{temp}}$, where the $i$-th query $Q_{\text{temp},i}$ is added to all tokens in $z_{\text{segs}, i}$. Furthermore, we introduce learnable concept queries $Q_{\text{concepts}}$, where the $t$-th query $Q_{\text{concepts},t}$ is added to the $t$-th token across all segment tokens $z_{\text{segs}} = \{z_{\text{segs},i}\}_{i=1}^N$.
 Taken together, we compute the adapted segment tokens for the $t$-th token of segment $S_i$ as:
 \begin{equation}
     Q_{\text{final}, i, t} = z_{\text{segs}, i, t} + Q_{\text{temp}, i} + Q_{\text{concepts}, t} .
 \end{equation}
We denote the full adapted segment token set as $Q_\text{final} = \{Q_{\text{final},i,t}\}_{i,t}$. 
 Similar to our $\mathcal{F}_{\text{CS}}$ function, we introduce learnable inter-segment queries $Q_{\text{inter}}$ to adapt the key frames tokens $z_{\text{key}}$ to be compatible with the video QFormer $\mathcal{F}_\text{VQT}$. 
We define our CV tokenizer as a weighted sum of the key frames tokens (to retain the global video context) and the repurposed video QFormer $\mathcal{F}_\text{VQT}$:
\small
 \begin{equation}\label{eq:weighted_sum}
     \mathcal{F}_\text{CV}(z_\text{segs} \ | \ z_\text{key}) = z_\text{key} + w \mathcal{F}_{\text{VQT}}( Q_{\text{final}}; \text{concat}\{ Q_{\text{video}}, z_{\text{key}} + Q_{\text{inter}}\}),
 \end{equation}
 \normalsize
 where $w$ is a learnable scalar. 

\subsection{Learning objective}\label{sec:learning-obj}
We define the learning objective for optimizing the parameters of the introduced tokenizer functions $\mathcal{F}_\text{CS}$ and $\mathcal{F}_\text{CV}$ and the global affine transformation $\phi_\text{inter}$ as predicting the high-level task labels of instructional videos spanning at least a few minutes. This objective is akin to summarizing the long videos concisely. Given the instruction-tuned nature of the pretrained vLLM, we convert the high-level task labels such as ``fix a car engine'' into the instruction format by manually crafting a set of query and response templates for training (see supplemental). Let $P$ be a question prompt for a given input video $V$ and $R$ its corresponding response comprising a sequence of $M$ words $R = \{\hat{l}_1, \cdots, \hat{l}_M \}$ (each word is represented as a one-hot vector). We minimize the cross-entropy loss:
\small
\begin{equation}
    \mathcal{L}(V, P, R) = -\sum_{j=1}^M \hat{l}_j\log p(l_j | \hat{l}_{<j}, V, P),
\end{equation}
\normalsize
where $p(l_j | \hat{l}_{<j}, V, P)$ denotes the probabilities for the $j$-th word given the preceding ground truth words $\hat{l}_{<j}$. 

%% file: 4_experiments.tex
\section{Experiments}

\begin{table*}[t]
\begin{center}
\setlength\tabcolsep{2pt}
\vspace{-3mm}
\resizebox{0.85\linewidth}{!}{
\begin{tabular}{|l|c|c|c|c|c|}
\hline
 Approach & Training & LLM & LLM architecture &  \# input frames & Top 1 Acc (\%) \\
\hline
\rowcolor{lightgray} Human accuracy (upper bound) & - & - & - & -  & 76.20\\
\rowcolor{lightgray} Language prior & - & Flan-T5-xl & Encoder-decoder & -  & 35.92\\
Random & - & - & - & - & 20.00 \\
VIOLET \cite{fu2023empirical} & - & Bert-Base \cite{su2019vl} & Encoder & 5  & 19.90\\
Frozen-BiLM \cite{yang2022zero}  & MLM & DeBERTa-V2-XLarge \cite{he2020deberta} & Encoder & 90 & 26.90\\
Video-Llama (finetuned) & Captioning & Llama-2 & Decoder & 32 & 28.36\\
mPLUG-Owl \cite{ye2023mplug} & Captioning & Llama & Decoder & 5 & 31.10\\
InternVideo \cite{wang2022internvideo} & Contrastive & CLIP & Encoder & 90  & 32.10\\
Video-Llama \cite{zhang2023video} & Captioning & Llama-2 & Decoder & 128  & 33.25\\
MovieChat \cite{song2023moviechat} & Captioning & Llama-2 & Decoder & 128  & \underline{33.49}\\
\modelabb (ours) & Captioning & Llama-2 & Decoder & 64 & \textbf{40.42}\\
\hline 
\end{tabular}}
\vspace{-2mm}
\caption{\textbf{Zero-shot long video question answering on EgoSchema benchmark.} For all models, we report the best results obtained across varying number of input frames. Our \modelabb approach outperforms the base Video-Llama model despite using much fewer frames. We also include the results for a strong language prior baseline as well as human performance (highlighted in gray).}
\label{tab:zero-shot-egoschema-eval}
\end{center}
\vspace{-7mm}
\end{table*}

\noindent\textbf{Datasets.} We train our \modelabb approach on a filtered subset of 250K videos from the HowTo100M instructional video dataset \cite{miech2019howto100m}. The filtered subset contains longer videos that span from four to over thirty minutes. Please see the supplemental for details on how we filter the training data. We evaluate our approach on two zero-shot long video question answering tasks -- the multiple choice format in EgoSchema \cite{mangalam2023egoschema} and procedure-understanding in Seed-Bench \cite{li2023seed}. Additionally, we evaluate on the task of short-term action recognition \cite{li2023seed} to analyze if the introduced CS and CV functions are detrimental to understanding short videos. Note that we report the best results across different numbers of frames.
\smallskip

\noindent\textbf{Implementation details.} We build our approach off the publicly available Video-LLama model \cite{zhang2023video} and train for 2 epochs on the final filtered subset of Howto100M. During evaluation, we compute the log-likelihood for each candidate answer and select the highest-scoring option for fair comparison \cite{brown2020language,li2023seed}. We provide further details about our training setup in the supplemental.

\begin{table}[t]
\begin{center}
\setlength\tabcolsep{2pt}
\resizebox{0.7\linewidth}{!}{
\begin{tabular}{|l|c|c|}
\hline
LLM & Architecture &  Top 1 Acc (\%) \\
\hline
Random & - & 20.00 \\
 GPT-J  & decoder-only  & 9.94\\
 GPT-Neo  & decoder-only  & 17.21\\
Vicuna & decoder-only  & 21.45\\
 Llama-2 & decoder-only  & 26.03\\
 Flan-T5-xl  & encoder-decoder & \textbf{35.92}\\
\hline 
\end{tabular}}
\vspace{-2mm}
\caption{\textbf{Zero-shot long video question answering on EgoSchema with language priors.} We observe that the language priors with different LLMs serve as strong baselines.}
\label{tab:language-prior-egoschema-eval}
\end{center}
\vspace{-8mm}
\end{table}

\begin{table*}[h]
\begin{center}
\vspace{-2mm}
\setlength\tabcolsep{2pt}
\resizebox{0.75\linewidth}{!}{
\begin{tabular}{|l|c|c|c|c|c|c|}
\hline
&  & & LLM &   & Procedure  &  Action  \\
 Approach & Training & LLM & architecture&  \# input frames & Understanding & Recognition \\
\hline
\rowcolor{lightgray} Language prior & - & Vicuna & Decoder-only & -  & 23.83 & 27.30\\
\rowcolor{lightgray} Language prior & - & Flan-T5 & Encoder-decoder & -  & 25.42 & 23.16\\
\rowcolor{lightgray} Language prior & - & Llama & Decoder-only &  - & 26.17 & 32.99\\
\rowcolor{lightgray} Language prior & - & Llama-2 & Decoder-only & - & 22.65 & 27.07\\
Random & - & - & - & - & 25.00 & 25.00\\
mPLUG-Owl \cite{ye2023mplug} & Captioning & Llama & Decoder-only & 32  & 26.51 &  26.72\\
VideoChat \cite{li2023videochat} & Captioning &Vicuna & Decoder-only  & 32  & 27.27 & 34.89 \\
Video-ChatGPT \cite{maaz2023video} & Captioning & Vicuna & Decoder-only & 32  & 21.14 & 27.59\\
Valley \cite{luo2023valley} & Captioning & Vicuna & Decoder-only & 32  & 20.72 & 31.26\\
Video-Llama-2 \cite{zhang2023video} & Captioning & Llama-2 & Decoder-only & 32  & 25.42 & 35.52\\
InstructBLIP \cite{instructblip} & Captioning & Flan-T5 & Encoder-decoder & 8  & 27.10 & 33.10\\
MovieChat \cite{song2023moviechat} & Captioning & Llama-2 & Decoder-only & 32 & 26.76 & 34.37\\
InstructBLIP Vicuna \cite{instructblip} & Captioning & Vicuna & Decoder-only & 8  & 23.07 & 34.48\\
VPGTrans \cite{zhang2023transfer} & Captioning & Flan-T5 & Encoder-decoder & 8  & \underline{31.88} & \underline{39.54}\\
\modelabb (ours) & Captioning & Llama-2 & Decoder-only & 64 & \textbf{35.91} & \textbf{41.26}\\
\hline 
\end{tabular}}
\vspace{-1mm}
\caption{\textbf{Zero-shot video question answering on Seed-Bench.} Compared to state-of-the-art mLLMs, our \modelabb approach improves the capability of the vLLM to not only understand long temporal context in procedure understanding but also to recognize short actions. We also compare to language prior baselines with different LLMs (highlighted in gray).}
\label{tab:zero-shot-seedbench-eval}
\end{center}
\vspace{-7mm}
\end{table*}

\begin{table}[h]
\small
\vspace{0mm}
\begin{center}
\resizebox{\linewidth}{!}{
\begin{tabular}{|l|c|c|c|}
\hline
  & EgoSchema & Procedure & Action \\
 Approach & Benchmark & Understanding & Recognition \\
\hline
Base & 33.25 & 26.68 & 35.52\\
Base + CS & 36.93 & 30.20 & 38.74\\
Base + CS + CV & \textbf{40.42} & \textbf{35.91} & \textbf{41.26}\\
\hline 
\end{tabular}}
\vspace{-1mm}
\caption{\textbf{Model ablations on the zero-shot evaluation benchmarks.} We ablate the effectiveness of different queries introduced in our \modelabb approach on all three evaluation tasks.}
\label{tab:model_ablations}
\end{center}
\vspace{-10pt}
\end{table}

\begin{table}[h]
\small
\begin{center}
\resizebox{\linewidth}{!}{
\begin{tabular}{|c|c|c|c|}
\hline
Keep & Condition on $z_\text{key}$ & Temporal concept & EgoSchema \\
$z_\text{key}$ output & in CS tokenizer & queries $Q_\text{temp}, Q_\text{concepts}$ & Benchmark \\
\hline
\cmark & \cmark & \cmark & 33.61 \\
\xmark & \xmark & \cmark & 39.12 \\
\xmark & \cmark & \xmark & 39.20 \\
\xmark & \cmark & \cmark & \textbf{40.42}\\
\hline 
\end{tabular}}
\caption{\textbf{Additional model ablations on the EgoSchema benchmark.} We include additional ablation experiments over adding temporal queries in our CS tokenizer function as well as retaining the learnable inter-segment queries as input into the LLM. We observe that global context conditioning and introducing learnable parameters are beneficial towards adapting pretrained vLLMs.}
\label{tab:additional_model_ablations_tokenizer}
\vspace{-10pt}
\end{center}
\end{table}

\begin{table}[h]
\begin{center}
\setlength\tabcolsep{2pt}
\resizebox{0.85\linewidth}{!}{
\begin{tabular}{|l|c|c|c|}
\hline
Approach & Aggregate pre-LLM & EgoSchema \\
\hline
Base & \xmark & 33.25 \\ 
Average & \cmark & 33.39 \\ 
Memory module (Moviechat) \cite{song2023moviechat} & \cmark & 34.62 \\ 
Concatenation & \xmark & 35.72 \\ 
\modelabb (ours) & \cmark & \textbf{40.33} \\ 
\hline 
\end{tabular}}
\vspace{-1mm}
\caption{\textbf{Comparisons between pre- and post-LLM temporal context aggregation.} We observe that naively encoding each video segment separately and concatenating the entire sequence of video tokens into the LLM performs worse than aggregating the video tokens \emph{before} passing them into the LLM.}
\label{tab:main-visual-agg-study}
\end{center}
\vspace{-10mm}
\end{table}



\subsection{Quantitative comparison to baselines}

Besides the tasks of long video question answering and procedure understanding on the EgoSchema and Seed-Bench benchmarks, we also evaluate our \modelabb model on short-term action recognition.

\noindent \textbf{EgoSchema evaluation.} We report the results of our zero-shot evaluation on the EgoSchema benchmark in Table~\ref{tab:zero-shot-egoschema-eval}. In addition to state-of-the-art vLLMs, we also compare our proposed \modelabb approach to language prior baselines that are not included in Mangalam \etal~\cite{mangalam2023egoschema}. The language prior baselines only use the questions and candidate answers for predictions. Please refer to the supplemental for more details on how we prompt these LLMs given a question and each candidate answer. Note that we also modify the questions and answers to replace ``C'' with ``the camera wearer'' so that the words used are more similar to the data used to pretrain these language models. To begin, we observe that the Flan-T5 \cite{chung2022scaling} language prior serves as a very strong baseline on this benchmark. Despite not relying on the input videos at all, this language prior baseline outperforms most of the state-of-the-art video-language models by a significant margin. In some cases, Frozen-BiLM and InternVideo have also been finetuned on QA datasets including How2QA \cite{li2020hero} and MSRVTT \cite{xu2016msr}. This finding suggests that existing vLLMs are not able to perform long-term temporal reasoning well although they have been trained on large-scale curated video and text data.

To better understand this finding, we also conduct an analysis of different state-of-the-art LLMs to determine their impact. In Table~\ref{tab:language-prior-egoschema-eval}, we see that the language prior accuracy varies greatly across the different LLMs. For example, the Flan-T5 model performs better than the LLama-2 model by approximately $9\%$. On the other end of the spectrum, a similarly-sized autoregressive GPT-J LLM with 6B parameters performs significantly worse than random. Given that the question and answer options in EgoSchema were generated using powerful LLMs (\eg, GPT4 \cite{openai2023gpt}, Bard \cite{Bard}, and Claude \cite{Claude}), we hypothesize that Flan-T5's accuracy on this task is due to having learned a representation that is similar to the LLMs used to generate the evaluation data.

While both Video-Llama and our approach use Llama-2 as the base LLM, we observe that Video-Llama still underperforms the Flan-T5 language prior baseline despite improving upon the Llama-2 language prior variant. In contrast, our \modelabb approach not only outperforms the Flan-T5 model, but also improves upon the base Video-Llama model by $\sim$7\%. \textbf{This finding demonstrates the effectiveness of our introduced tokenizer functions at reasoning about temporal relations over longer spans.} One question that arises from these results is whether the accuracy gains by \modelabb can be attributed to further training on video data that may be semantically similar to the target domain. To address this question, we also finetune the Video-Llama captioning model without our CS and CV functions. Finetuning Video-LLama yields a drop of $\sim$5\% in top-1 accuracy from the base Video-Llama model, and suggests that the improvements are not solely due to further finetuning. We include details about finetuning Video-Llama on HowTo100M in the supplemental.

\noindent \textbf{Seed-Bench Procedure Understanding.} We report the results of our evaluations on the procedure understanding task of the Seed-Bench benchmark in Table~\ref{tab:zero-shot-seedbench-eval}. The goal of procedure understanding is to detect all actions performed in a given video and arrange them in the correct temporal order, which requires fine-grained temporal understanding over a long span. As shown in Li \etal \cite{li2023seed}, state-of-the-art vLLMs (\eg, mPLUG-Owl, VideoChat, and Video-Llama) often perform worse than their image-based variants such as InstructBLIP and VPGTrans. In certain cases, some vLLMs actually perform worse than their base LLM language prior baselines. For instance, using videos causes the accuracy to drop by 2-3\% in the case of Valley \cite{luo2023valley} and Video-ChatGPT \cite{maaz2023video} when compared to their base Vicuna LLM \cite{vicuna2023} language prior.

It is also notable that large-scale pretraining on millions of short video and caption pairs only helps Video-Llama to improve by $\sim$4\% over its base Llama-2 language prior. This finding suggests that learning to aggregate temporal context over a larger number of key frames without knowledge of the global context does not result in learning an effective key frames tokenizer function. In contrast, we observe that our proposed \modelabb model gains an improvement of $\sim$9\% over Video-Llama in spite of the lightweight finetuning stage that uses many fewer training videos as compared to the initial pretraining on WebVid10M \cite{Bain21} and curated instructional video data \cite{maaz2023video}. This finding suggests that our introduced CS and CV tokenizer functions are beneficial towards reasoning about long-term temporal relations between different action sequences in videos.

\noindent \textbf{Seed-Bench Action Recognition.} Finally, we evaluate our \modelabb approach on the task of action recognition (Table~\ref{tab:zero-shot-seedbench-eval}) to study the effect of our introduced tokenizer functions for short-term temporal understanding tasks. In contrast to the longer setting in the procedure understanding task, the videos in this task generally have duration of around 10 seconds. Similar to our observations on the procedure understanding task, the mPLUG-Owl, Video-ChatGPT, and Valley vLLMs perform worse on this task than the image-based InstructBLIP and VPGTrans models. 

Note that the base Video-Llama model performs worse than the image-LLM VPGTrans by $\sim$4\% despite its large-scale pretraining on seconds-long videos. This finding suggests that its key frames tokenizer function may be limited at reasoning about fine-grained actions and interactions between objects. While we are primarily focused on understanding long videos, we observe that our CS and CV tokenizer functions are also beneficial to understanding short actions, improving upon Video-Llama by $\sim$6\% and outperforming VPGTrans by $\sim$2\%. These results suggest that using key frames to provide global context for reasoning about spatiotemporal relationships between video segments may be crucial for fine-grained action understanding. 

\begin{figure*}[t!]
    \subfloat[]{%
        \includegraphics[width=.5\linewidth]{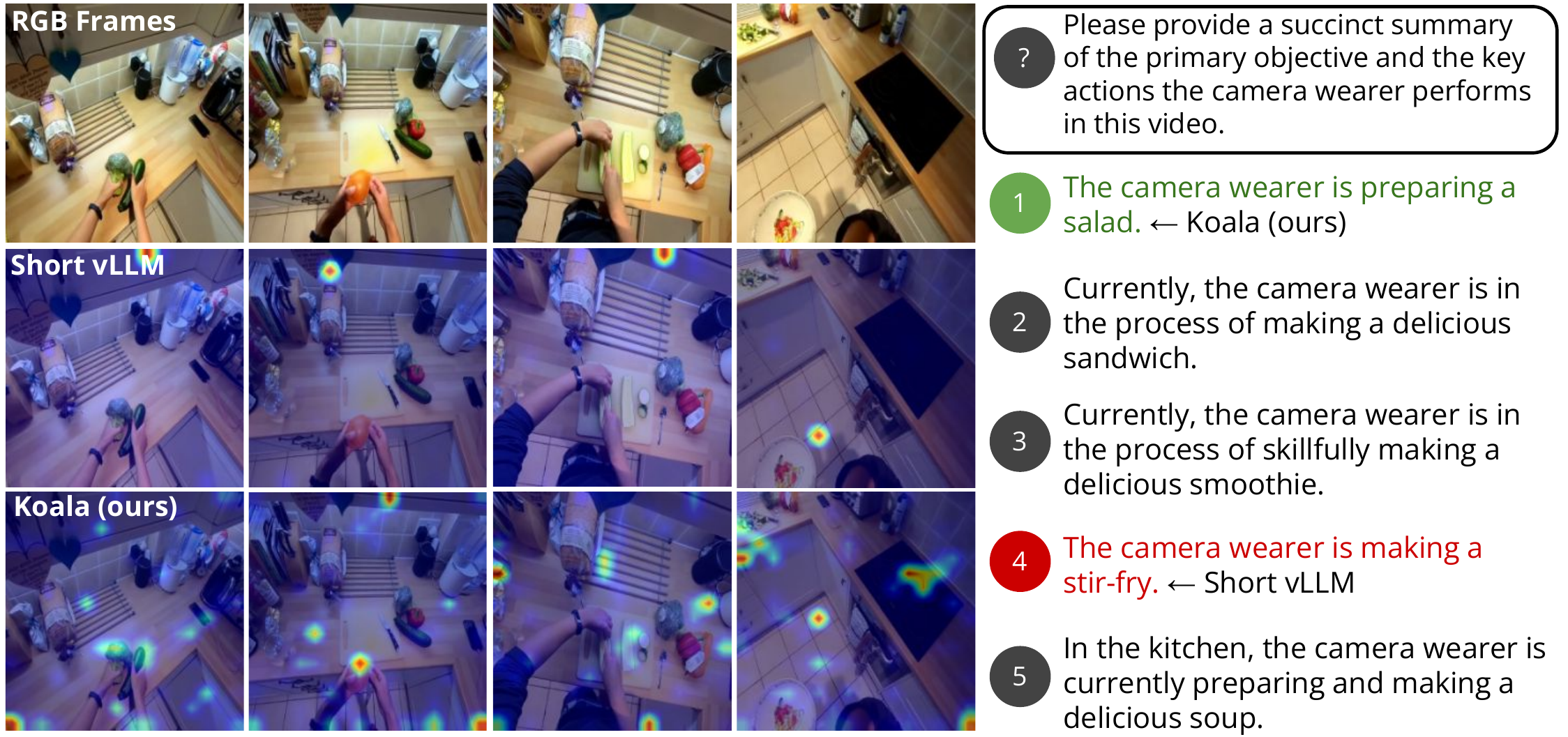}%
        \label{subfig:attn_map_first}%
    }\hfill
    \subfloat[]{%
        \includegraphics[width=.5\linewidth]{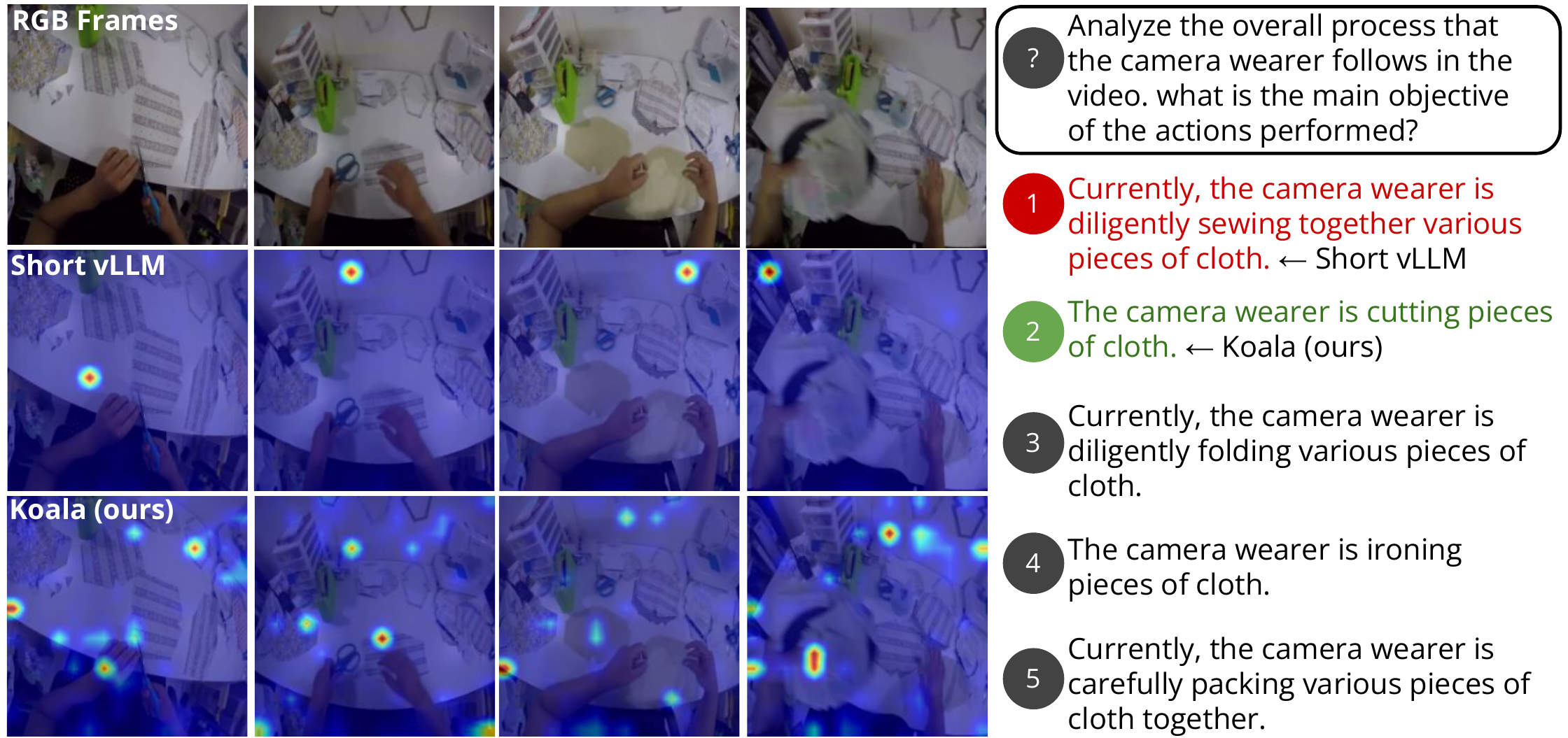}%
        \label{subfig:attn_map_second}%
    }
    \vspace{-10pt}
    \caption{\textbf{Example attention heatmap visualizations on EgoSchema.} We provide some qualitative examples of predictions made by our \modelabb approach and the base Video-Llama model based on what they focus on. We observe that \modelabb is generally able to focus on relevant regions better than the base vLLM.}
    \label{fig:main-egoschema_attn_maps}
\end{figure*}

\subsection{Ablation study}\label{sec:ablations}
\noindent \textbf{Overall \modelabb architecture.} In Table~\ref{tab:model_ablations}, we ablate our CS and CV functions across all three evaluation tasks to determine their individual contributions. Consistent across all three tasks, we observe that conditioning on the key frames for global context to aggregate spatiotemporal context within each video segment in our CS function is especially crucial, as evidenced by a $\sim$3\% improvement in top-1 accuracy on average. We also note the importance of reasoning about spatiotemporal contextual information between segments in our CIS function where our concept queries help improve accuracy on both long and short-term temporal understanding. 

\smallskip
\noindent \textbf{Tokenizer design.} We ablate the design choices of the CS and CV tokenizers on the EgoSchema benchmark in Table~\ref{tab:additional_model_ablations_tokenizer}. We observe that passing the output tokens corresponding to $z_\text{key}$ into the LLM (``Keep $z_\text{key}$ output'') instead of discarding them leads to a sharp drop in accuracy of $\sim$7\%, which may be due to the base vLLM being pretrained to accept a fixed number of video tokens as input. Additionally, we note the benefit of conditioning the CS tokenizer on the key frame tokens $z_\text{key}$, where the lack of conditioning leads to a drop of $\sim$1.3\%. Finally, we observe the importance of introducing additional parameters in the form of the temporal concept queries $Q_\text{temp}$ and $Q_\text{concepts}$ in the CV tokenizer. As evidenced by the accuracy gain, it is important to adapt to the frozen video QFormer $\mathcal{F}_\text{VQT}$.

\noindent \textbf{Temporal aggregation.} Lastly, given the recent importance of vLLMs, we study in Table~\ref{tab:main-visual-agg-study} the key factors for integrating long-term temporal context from more input frames into the frozen vLLM and compare to our \modelabb. For all aggregation function variants, we use 4 segments of 8 frames each. We next describe the different aggregation variants. The first variant (``Average'') obtains visual tokens by averaging $\frac{1}{N}\sum_{i=1}^N\mathcal{F}_\text{key}(S_i)$ across all video segments $S_i$. These averaged tokens are concatenated with the key frame tokens $z_\text{key}$ before being projected by $\phi_\text{key}$ and passed to the LLM. The second variant (``Memory module'') utilizes short and long-term memory mechanisms \cite{song2023moviechat} to compute contextualized soft visual tokens as input into the LLM. We pass in $\mathcal{F}_\text{key}(S_i)$ across all video segments into the short-term memory and use the long-term memory tokens as input into the LLM. In the third variant (``Concatenation''), we concatenate tokens $\mathcal{F}_\text{key}(S_i)$ across all segments $S_i$, allowing the LLM to leverage its pretrained self-attention function for temporal reasoning. We note that this variant is similar to the SlowFast approach \cite{feichtenhofer2019slowfast}, where the “slow” frame features $z_\text{key}$
are fused with the “fast” frame features $\mathcal{F}_\text{key}(S_i)$ by concatenation.





In general, we observe that it is more beneficial to aggregate temporal context in videos and encode it in the sequence of visual tokens before passing them into the LLM. While average-pooling video segment representations or using a long-term memory module \cite{song2023moviechat} may lose some fine-grained spatiotemporal information, we observe that they are outperformed by the concatenation variant on downstream evaluations by only a small margin. This finding suggests that the self-attention layers in the LLM may not understand longer sequences of visual tokens without additional large-scale pretraining. Finally, we further ablate over the training hyperparameters including the number of segments as well as frames per segment used as input into the vLLM. Please refer to the supplemental for these results.

\subsection{Qualitative results}
We analyze how our introduced spatiotemporal queries in the CS and CV tokenizer functions change what the vLLM focuses on in the input videos (Figure~\ref{fig:main-egoschema_attn_maps}). Compared to the baseline Video-Llama model, we observe that our introduced queries generally help to improve the capability of the model to focus on relevant visual concepts. The visualization in Figure~\ref{fig:main-egoschema_attn_maps}\textcolor{red}{a} is particularly interesting because the introduced queries help our \modelabb model to predict that the person is making a salad based on its attention on the empty stove in the last frame (far right). Additionally, we also observe in Figure~\ref{fig:main-egoschema_attn_maps}\textcolor{red}{b} that our model generally focuses on the pieces of cloth as opposed to the background as in the case of the base Video-Llama model.

\noindent \textbf{Limitations.} While our \modelabb approach is able to extend the video tokenizer function of a pretrained vLLM to understand minutes-long videos, it may still be limited at understanding much longer videos such as movies. Since it relies on a pretrained model, we inherit as a fundamental limitation a maximum number of input tokens, thereby limiting the number of input segments. However, extending positional embeddings to longer sequences remains an open work, especially in the setting of vLLMs.


%% file: 5_conclusion.tex
\section{Conclusion}
In conclusion, we propose an approach, \modelabb, that introduces the Conditioned Segment and Conditioned Video tokenizer functions. Our CS and CV functions leverage learnable spatiotemporal queries to adapt the frozen video tokenizer function in pretrained vLLMs to generalize to minutes-long videos. More importantly, we empirically demonstrate the benefits of our \modelabb approach where it improves the base vLLMs on both short and long-term temporal understanding tasks. 

\noindent 
\textbf{Acknowledgements}: This material is based upon work supported, in part, by DARPA under agreement number HR00112020054. 

%% file: 6_supplemental.tex

\begin{table*}[t!]
\small
\begin{center}
\begin{tabular}{l|l}
\hline
\textbf{Prompt template} & \textbf{Response template} \\
\hline
 $<$VISUAL$>$ What is the most likely objective in the video? [/INST] & The most likely objective in the video is to \{task label\}. \\

$<$VISUAL$>$ What is the most likely goal in the video? [/INST]& The most likely goal is to \{task label\}. \\

$<$VISUAL$>$ What is the person trying to do in the video? [/INST] & The person is trying to \{task label\}. \\

$<$VISUAL$>$ What is happening in the video? [/INST]& This video demonstrates the steps to \{task label\}. \\

$<$VISUAL$>$ Describe the most likely objective in the video. [/INST] & The most likely objective in the video is to \{task label\}. \\

$<$VISUAL$>$ Describe the most likely goal in the video. [/INST]& The most likely goal is to \{task label\}. \\

$<$VISUAL$>$ Describe what the person is trying to do in the video. [/INST] & The person is trying to \{task label\}. \\

$<$VISUAL$>$ Describe what is happening in the video. [/INST] & This video demonstrates the steps to \{task label\}. \\
\hline 
\end{tabular}
\caption{\textbf{Instruction and sample response templates.} We use these templates to transform high-level goal labels of the finetuning dataset into the instruction tuning format during our finetuning stage. We use $<$VISUAL$>$ as a placeholder for the expression $[\text{INST}]<$Video$><$ImageHere$><$/Video$>$. Note that we substitute the $<$ImageHere$>$ token with the final contextualized video tokens in practice during finetuning and downstream evaluations.}
\label{tab:instruction-response-templates}
\end{center}
\end{table*}

In this supplemental, we provide the following additional material to the main paper:
\begin{enumerate}
    \item[A] Manually crafted query and response templates
    \item[B] CLIP filtering process for HowTo100M
        \begin{enumerate}
            \item CLIP score filtering
            \item Qualitative visualizations 
        \end{enumerate}
    \item[C] Implementation details for training and evaluation
    \item[D] Evaluation benchmark details
        \begin{enumerate}
            \item EgoSchema
            \item Seed-Bench Procedure Understanding
            \item Seed-Bench Action Recognition
        \end{enumerate}
    \item[E] Additional evaluations on the NExT-QA benchmark
    \item[F] Additional ablation experiments
        \begin{enumerate}
            \item Baseline model definitions 
            \item Efficiency of aggregating temporal context in videos pre-LLM 
            \item Ablation over training hyperparameters
        \end{enumerate}
    \item[G] Additional qualitative visualizations
\end{enumerate}

\begin{figure*}[t!]
\captionsetup[subfigure]{labelformat=empty}
  \centering
  \subfloat[a][]{\includegraphics[width=0.95\linewidth]{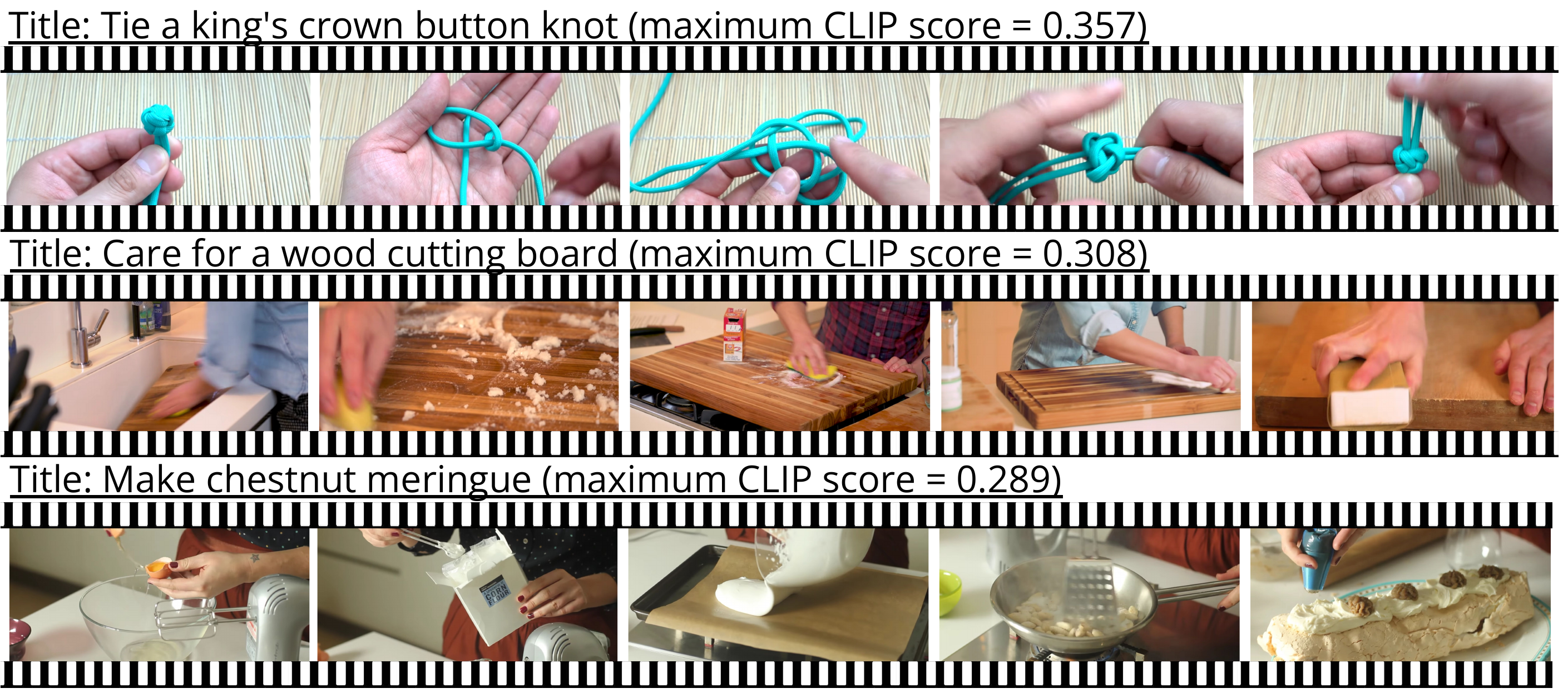}} \\
  \vspace{-15pt}
  \subfloat[b][]{\includegraphics[width=0.95\linewidth]{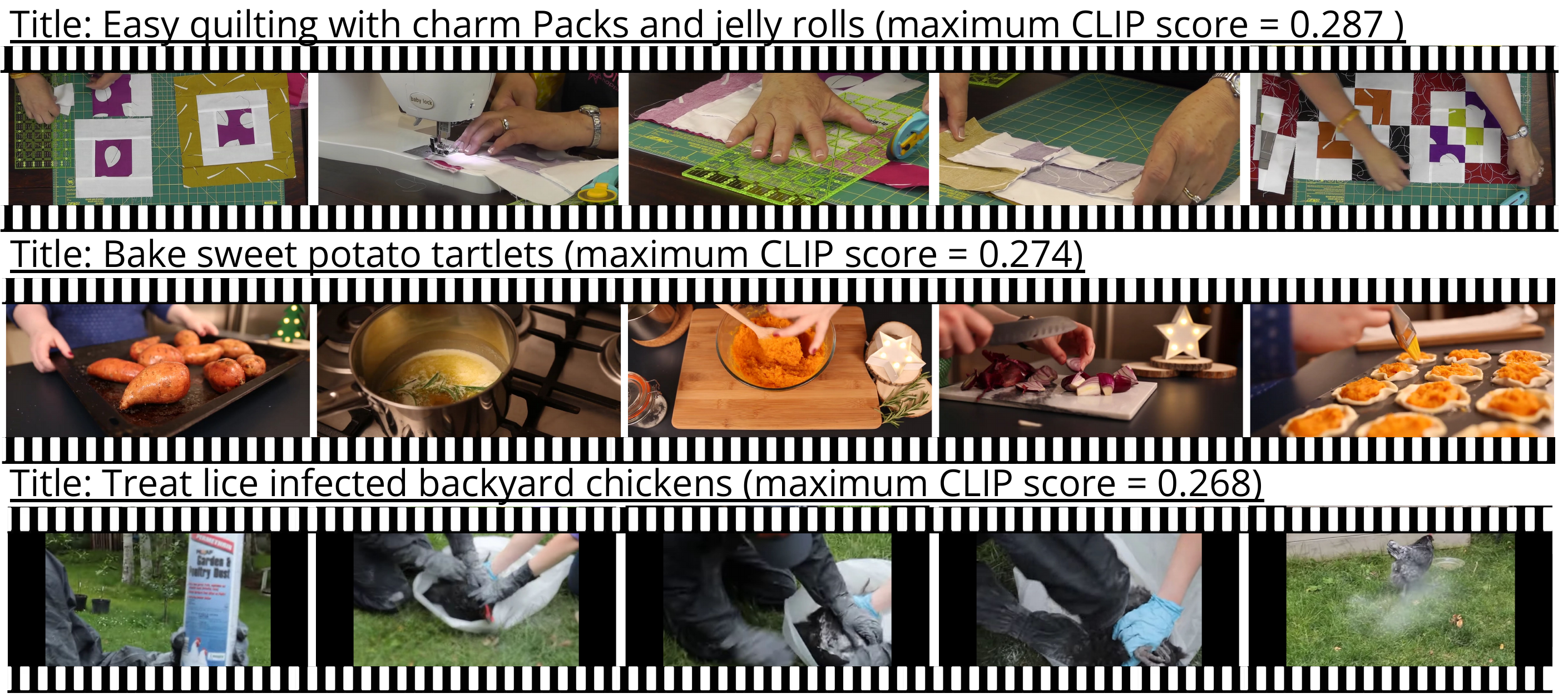}}
  \caption{\textbf{Examples of videos filtered using maximum CLIP scores of video frames with respect to their task labels.} We use the
CLIP [48] model to compute a similarity score between each extracted frame and the corresponding task label of the video. We generally
observe that filtering videos based on the maximum CLIP score of any frame with respect to the task label results in videos with more
visual diversity.} \label{fig:clip_filtering_comparisons}
\end{figure*}

\section{Instruction templates}
As mentioned in the main paper, we train our Koala approach on instructional videos from the HowTo100M dataset \cite{miech2019howto100m}. The videos are sourced from YouTube using a list of high-level activities obtained from WikiHow\footnote{https://www.wikihow.com/}. As such, each instructional video has a corresponding high-level task label such as ``replace a car tire'' and ``make a bacon lettuce and tomato sandwich.'' Given the instruction-tuned nature of the base video-LLM, we manually craft question and response templates as shown in Table~\ref{tab:instruction-response-templates}. In Table~\ref{tab:instruction-response-templates}, we use $<$VISUAL$>$ as a placeholder for the expression ``$[\text{INST}]<$Video$><$ImageHere$><$/Video$>$.'' During finetuning and downstream evaluations, we substitute the ``$<$ImageHere$>$'' token with the final contextualized video tokens and substitute ``\{task label\}'' with the corresponding high-level task label. For training, we create the question prompt $P$ and response $R$ by randomly sampling a pair from Table~\ref{tab:instruction-response-templates}. 

\section{CLIP filtering of training data}
We observe instances where the high-level task labels are not visually relevant to the video content. An example of the aforementioned instances is a video of a person simply describing an action without showing it. Given the demonstrated importance of clean data \cite{gadre2023datacomp} in training instruction-tuned foundation models, we perform video filtering using the pretrained CLIP ViT-L14 \cite{radford2021learning} model variant. 

Specifically, we use CLIP's visual and text encoders $\text{CLIP}_{\text{visual}}$ and $\text{CLIP}_{\text{text}}$ to measure the similarity between $N$ encoded extracted frames for each video $V = \{V_i\}_{i=1}^N$ and its corresponding task label $L$. We uniformly sample 128 frames from each video and keep the video if it satisfies the following constraint:
\begin{equation}
    \max\limits_{V_i\in V}(\text{CLIP}_{\text{visual}}(V_i)^T \text{CLIP}_{\text{text}}(L)) \geq \tau,
\end{equation}
where $\tau$ denotes the cosine similarity threshold.

We show examples of filtered videos using the maximum CLIP scores in Figure~\ref{fig:clip_filtering_comparisons}. In the filtering process, we generally observe that selecting videos based on the maximum relevance score of any frame with respect to the high-level task labels yields videos with increased visual diversity across its frames, as compared to using the mean score across all sampled frames. We set $\tau$ to be 0.26 in practice after manually inspecting the visual relevance of about 500 videos and their corresponding similarity scores between the video frames and the corresponding task label.

\section{Implementation details}
\noindent\textbf{Training.} We optimize the learnable weights of our introduced Conditioned Segment (CS) and Conditioned Video (CV) functions using the AdamW \cite{loshchilov2017decoupled} optimizer for two epochs. We implement our model by building on the LVIS library \cite{li-etal-2023-lavis}. We also adopt a linear warmup schedule over 10\% of training steps with a maximum learning rate of $1e^{-5}$ and gradually anneal it based on a cosine schedule. Our final filtered training set consists of approximately 250K videos in total. In this work, we build our approach off the state-of-the-art Video-LLama \cite{zhang2023video} model. We train our model on 4 RTX 6000 GPUs. We also define the dimensionality of the outputs of key frames, contextualized segment and inter-segment tokens.  For a set of $T$ key frames $V_\text{key}$, we define the output of the key frames tokenizer function $\mathcal{F}_\text{key}$ as: $z_\text{key} \in \mathbb{R}^{N \times D}$, where $N$ and $D$ denote the number and dimensionality of the frozen video queries $Q_\text{video}$, respectively. The outputs of our Conditioned Segment and Conditioned Video tokenizer functions $z_\text{segs}$ and $z_\text{inter}$ also have similar dimensionality of $\mathbb{R}^{N \times D}$. 

Similarly, our segment and inter-segment queries have the same dimensionality of $\mathbb{R}^{N \times D}$. The LLM linear projection functions $\phi$ project the dimensionality of the key frames tokens $z_\text{key}$ and contextualized inter-segment tokens $z_\text{inter}$ from $D$ to $D^f$ where $D^f$ denotes the dimensionality of the textual tokens as input into the frozen LLM. Similar to prior work \cite{zhu2023minigpt, zhang2023video}, we set $N$, $D$ and $D^f$ to be 32, 768 and 4096, respectively. The final value of $w$ in Equation \textcolor{red}{6} (main) is 0.0203.

\textbf{Downstream evaluations.} 
We adopt the same evaluation method of calculating log-likelihood for each candidate answer and selecting the highest-scoring option for fair comparisons with prior work \cite{brown2020language,li2023seed}. Note that we include the soft video tokens (Section \textcolor{red}{3} main) in all question-answer prompts. Given the instruction-tuned and generative nature of our final vLLM, we formulate an input text prompt for the zero-shot evaluations on the downstream multiple-choice question answering benchmarks. Specifically, for each question $Q$ and the set of answer options $A = \{a_1, \cdot\cdot\cdot, a_{||A||} \}$, we experiment with the following manually-crafted text prompt for the $j$-th candidate answer $a_j$: \textcolor{MidnightBlue}{``Given the question $<$Q$>$, the answer is $<$$a_j$$>$.''} We compute the final prediction for each question by selecting the answer option that returns the highest logit score for the question and candidate answer pair. For all models and evaluation datasets, we report the best results obtained across varying number of input frames.

\section{Evaluation datasets}
\textbf{Zero-shot evaluation benchmarks.} Our main goal is to introduce an approach for long-form video understanding. Consequently, we evaluate our proposed \modelabb approach on several zero-shot long video question answering tasks with the multiple choice format including EgoSchema \cite{mangalam2023egoschema} and procedure-understanding in Seed-Bench \cite{li2023seed}. Additionally, we also evaluate on the task of  short-term action recognition \cite{li2023seed} to analyze if the introduced CS and CV functions are detrimental to understanding short videos.
\begin{enumerate}
    \item EgoSchema \cite{mangalam2023egoschema} - EgoSchema is a challenging long video question-answering benchmark that contains 5031 3-minutes long videos and each question contains 5 possible options.
    
    \item Seed-Bench Procedure Understanding \cite{li2023seed} - The procedure understanding task contains 1170 questions with 4 answer options and the goal is to select the option that specifies the correct sequence of actions.
    
    \item Seed-Bench Action Recognition \cite{li2023seed} - To determine the effectiveness of \modelabb on short-term temporal understanding, we also evaluate on the action recognition task, which contains 1740 questions.

    \item NExT-QA \cite{xiao2021next} - The NExT-QA dataset evaluates a video model's capability to describe and explain temporal actions in videos. NExT-QA contains approximately 52K question-answer pairs for 5,440 videos. Additionally, these questions are split into several categories such as temporal or descriptive.
\end{enumerate}

\section{Additional evaluations}
\begin{table}[h]
\begin{center}
\resizebox{\linewidth}{!}{
\begin{tabular}{|l | c c c c|}
\hline
Video-LLM & $\text{Acc}_\text{C}$ & $\text{Acc}_\text{T}$ & $\text{Acc}_\text{D}$ & $\text{Acc}_\text{AVG}$\\
\hline
Video-Llama (finetuned) & 27.43 & 32.14 & 32.38 & 29.71\\
VideoLlama & 31.32 & 35.49 & \textbf{42.64} & 34.47\\
MovieChat & 31.12 & 35.80 & 42.49 & 34.43\\
Koala (ours) & \textbf{32.83} & \textbf{38.13} & 41.21 & \textbf{35.85}\\
\hline 
\end{tabular}}
\caption{\textbf{Zero-shot evaluation on NExT-QA test split.} We observe that our Koala model performs better than other approaches across most of the different video understanding tasks.}
\label{tab:nextqa-eval}
\end{center}
\end{table}
We report the results of our zero-shot evaluation on the test split of the NExT-QA [{\color{green} A}] benchmark in Table~\ref{tab:nextqa-eval}. NExT-QA divides its questions into three categories: \textbf{(1) Causal (C)}, \textbf{(2) Temporal (T)}, \textbf{(3) Description (D)}. Compared to prior work, our approach achieves higher accuracy across the \textbf{Causal (C)} and \textbf{Temporal (T)} categories, demonstrating its effectiveness at understanding long temporal context. However, our approach under-performs on \textbf{Description (D)} questions that involve counting the ordinality of objects. This result suggests that using curated descriptive annotations for the final finetuning stage, as done in prior work [{\color{green} 32}, {\color{green} 46}, {\color{green} 64}], may be beneficial for understanding such concepts.

\section{Ablation model baselines and efficiency metrics}
We provide additional implementation details on the baseline models in Section \textcolor{red}{4.2} of the main paper here before describing their performance and efficiency trade-offs. Recall that our goal is to compare Koala to these baselines to better understand how to integrate long-term temporal visual context with vLLMs.

\noindent \textbf{Average.} In contrast to existing vLLMs which often just extract a small and fixed number of key frames for each video regardless of its temporal duration, we subsample $S$ segments of $T$ key frames. We encode each segment separately with the key frames tokenizer $\mathcal{F}_\text{key}$ and average-pool the key frames tokens over the segments to compute the final visual input $z_\text{final}$ into the base LLM.  Specifically, we compute the final input as:
\begin{equation}
    z_\text{final} = \frac{1}{N} \sum_{i=1}^N \mathcal{F}_\text{key}(S_i),
\end{equation}
where $S_i$ denotes the frames for the $i$-th segment.

\begin{figure*}[t]
\begin{center}
\includegraphics[width=1.0\linewidth]{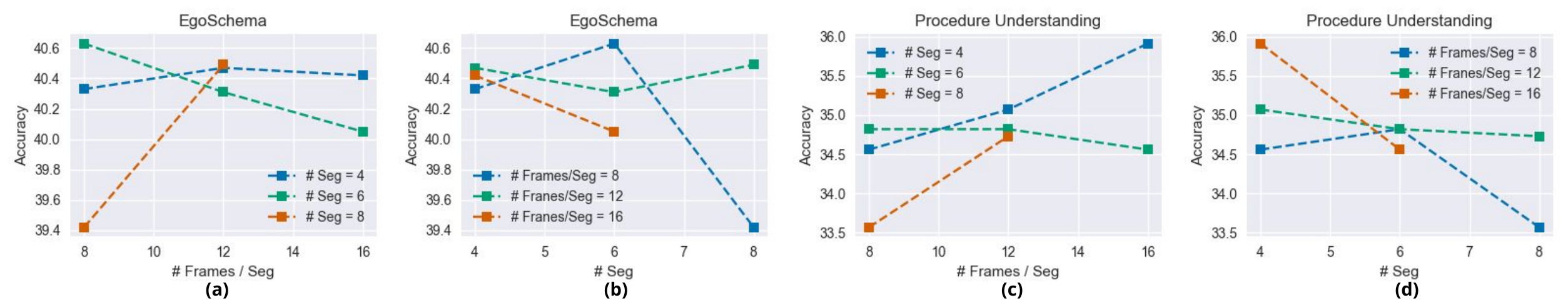}
\end{center}
   \caption{\textbf{Ablation over number of segments and frames.} Increasing the number of frames per segment while using a smaller number of segments during training is generally beneficial for long video understanding. We note that we run into an out-of-memory error with 8 segments of 16 frames each.} 
\label{fig:num_frames_segments_ablation}
\end{figure*} 

\begin{table}[t!]
\begin{center}
\setlength\tabcolsep{2pt}
\begin{tabular}{|l|c|c|c|c|}
\hline
 & Aggregate & & \\
Approach & pre-LLM & EgoSchema & GFLOPs \\
\hline
Base & No & 33.25 & 12K \\ 
Average & Yes & 33.39 & 12K \\ 
Memory module \cite{song2023moviechat} & Yes & 34.62 &  12K\\ 
Concatenation & No & 35.72 & 15K \\ 
\modelabb (ours) & Yes & \textbf{40.33} & 13K \\ 
\hline 
\end{tabular}
\caption{\textbf{Comparison of performance and efficiency tradeoffs between different video aggregation baselines.} We observe that our \modelabb approach improves the ability of the base vLLM for long-term temporal understanding significantly while only increasing the computational cost marginally.}
\label{tab:visual-agg-study}
\end{center}
\end{table}

\noindent \textbf{Memory module.} A common approach to model long-term temporal context for long videos is to use a feature memory module to store representations of past video segments for lookup. Inspired by \cite{song2023moviechat, cheng2022xmem}, we also adopt a simple baseline using a short-term memory module as well as a long-term memory module to mitigate the issue of forgetting information from the distant past. At a high level, we
pass in $\mathcal{F}_\text{key}(S_i)$ across all video segments into the short-
term memory and use the long-term memory tokens as input into the LLM. 

The key frames tokenizer function in pretrained vLLMs is often limited by the maximum number of key frames that can be used as input due to the length of the sequence of learnt temporal positional embeddings. 
To extend the original sequence of positional embeddings, we adopt an approach \cite{su2020pos} to hierarchically decompose the learnt positional embeddings such that we can extend them from its initial length $n$ to $n^2$. We refer interested readers to Song \etal \cite{song2023moviechat} for more details.

\noindent \textbf{Concatenation.} Last but not least, we also introduce the concatenation ablation to study the importance of aggregating temporal context over the input frames and encoding the information in the soft video tokens \emph{before} projecting them into the feature space of the base LLM. The concatenation baseline differs from the other baselines since it is relying on the self-attention layers in the pretrained LLM to aggregate temporal context over multiple segments of key frames. For this ablation, we encode each segment separately with $\mathcal{F}_\text{key}$ and concatenate the visual tokens from all segments as input into the LLM instead of average-pooling them. Mathematically, we formalize this operation as such:
\begin{equation}
    z_\text{final} = \text{concat}\{\mathcal{F}_\text{key}(S_1), \cdot \cdot \cdot, \mathcal{F}_\text{key}(S_N)\},
\end{equation}
where $\text{concat}\{ \}$ denotes the concatenation operation.

\noindent\textbf{Trade-off between performance and efficiency.} In addition to the performance on the EgoSchema benchmark, we also compare the performance and efficiency trade-offs between the different baselines in Table~\ref{tab:visual-agg-study}. We observe that the concatenation baseline not only performs worse at understanding long videos but is also the most computationally expensive variant with 15K GFLOPS. This is reasonable since we are computing the full self-attention operation over the extended sequence of video tokens in each layer of the base LLM. In contrast, while our \modelabb approach uses $\sim$1K GFLOPS more than the base, average and memory module baselines, it outperforms them by a significant margin of $\sim$6\%.

\smallskip
\noindent\textbf{Ablation over number of segments and frames per segment.} In Figure~\ref{fig:num_frames_segments_ablation}, we study the effect of varying the number of video segments and frames within each segment during training. In general, we observe that increasing the number of frames per segment (Figure~\ref{fig:num_frames_segments_ablation}\textcolor{red}{a} and \textcolor{red}{c}) while reducing the number of segments (Figure~\ref{fig:num_frames_segments_ablation}\textcolor{red}{b} and \textcolor{red}{d}) is generally beneficial for long video understanding, as exemplified by the $\sim$1.5\% increase in accuracy on procedure understanding when the number of frames per segment increases from 8 to 16 with 4 segments. The drop in accuracy with increasing segments may be due to redundant information factored into the temporal context aggregation.

\section{Additional qualitative visualizations}

\begin{figure*}[!t]
\captionsetup[subfigure]{labelformat=empty}
  \centering
  \subfloat[c][(c) Example prediction 3]{\includegraphics[width=\linewidth]{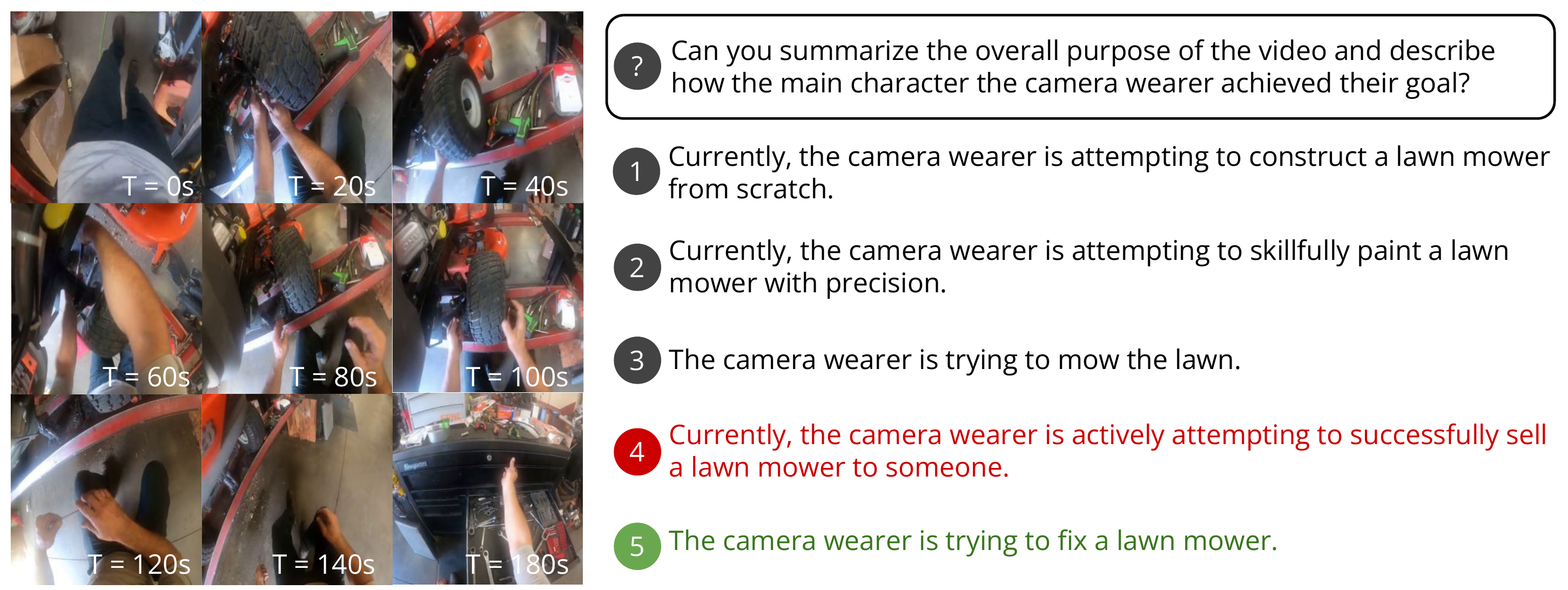}}

  \subfloat[d][(d) Example prediction 4]{\includegraphics[width=\linewidth]{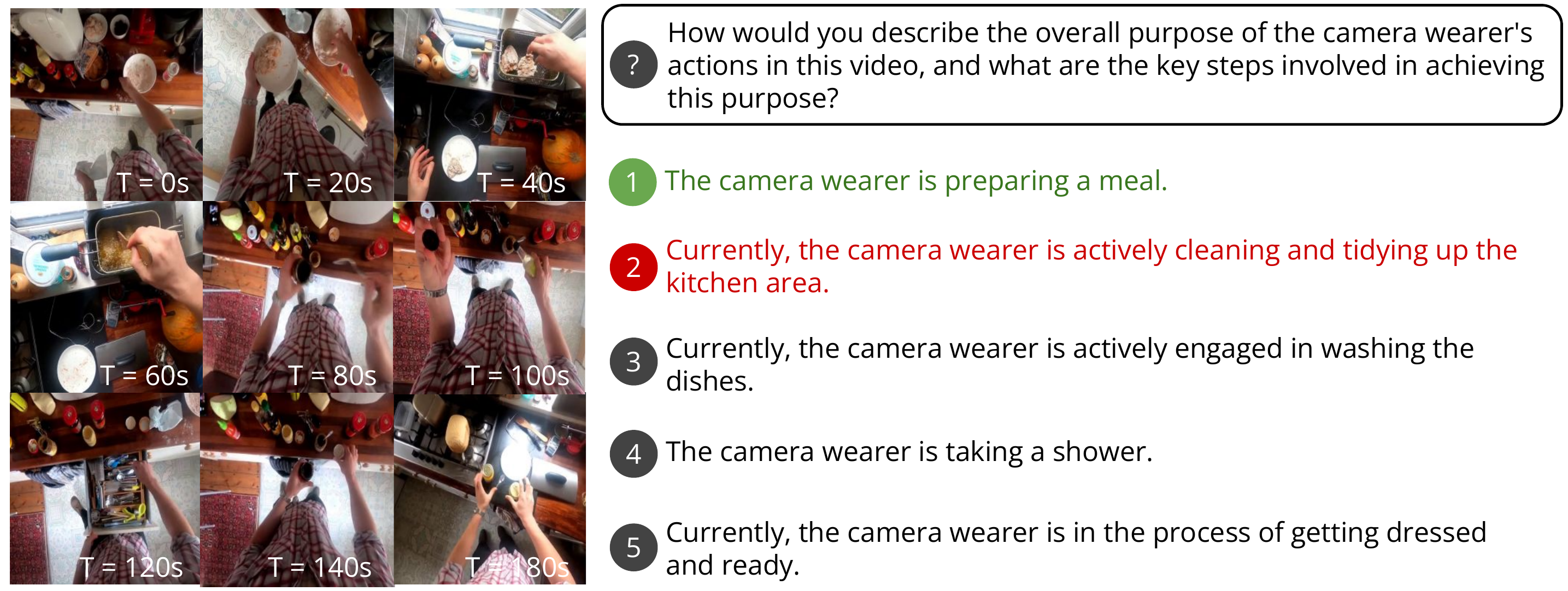}}
  \caption{\textbf{Sample predictions on EgoSchema.} We provide some qualitative examples of predictions made by our proposed Koala approach and the base Video-Llama model on the very challenging long-term video understanding EgoSchema benchmark.} \label{fig:egoschema_pred_samples}
\end{figure*}

\noindent\textbf{Visual examples of EgoSchema predictions.} To gain insights into how our introduced spatiotemporal queries have helped improve the long-term temporal understanding capability of the frozen base vLLM, we provide several examples of correct predictions on the very challenging EgoSchema benchmark in Figure~\ref{fig:egoschema_pred_samples}. Note that while EgoSchema is meant as a zero-shot evaluation benchmark, we use the subset of evaluation samples for which the correct answers are provided in these visualizations. 

In Figures~\ref{fig:egoschema_pred_samples}\textcolor{red}{a} and \ref{fig:egoschema_pred_samples}\textcolor{red}{b}, we see that the model often makes its predictions based on the first few input video frames and does not incorporate visual information from the entire videos, resulting in limited temporal context. In contrast, our approach is able to incorporate information over a larger time window, allowing it to summarize videos more accurately. Additionally, we also see using the spatiotemporal queries also encourage the base vLLM to hallucinate less visual details (Figures~\ref{fig:egoschema_pred_samples}\textcolor{red}{c} and \ref{fig:egoschema_pred_samples}\textcolor{red}{d}), resulting in more accurate summarizations. Since it may be a little difficult to understand minutes-long videos from just a few select key frames, we have also attached the videos as part of the supplemental submission for reference. 

\begin{figure*}[t!]
  \centering
  \subfloat[a][Video link: \url{https://www.youtube.com/watch?v=TvmFKsmatbI}]{\includegraphics[width=0.95\linewidth]{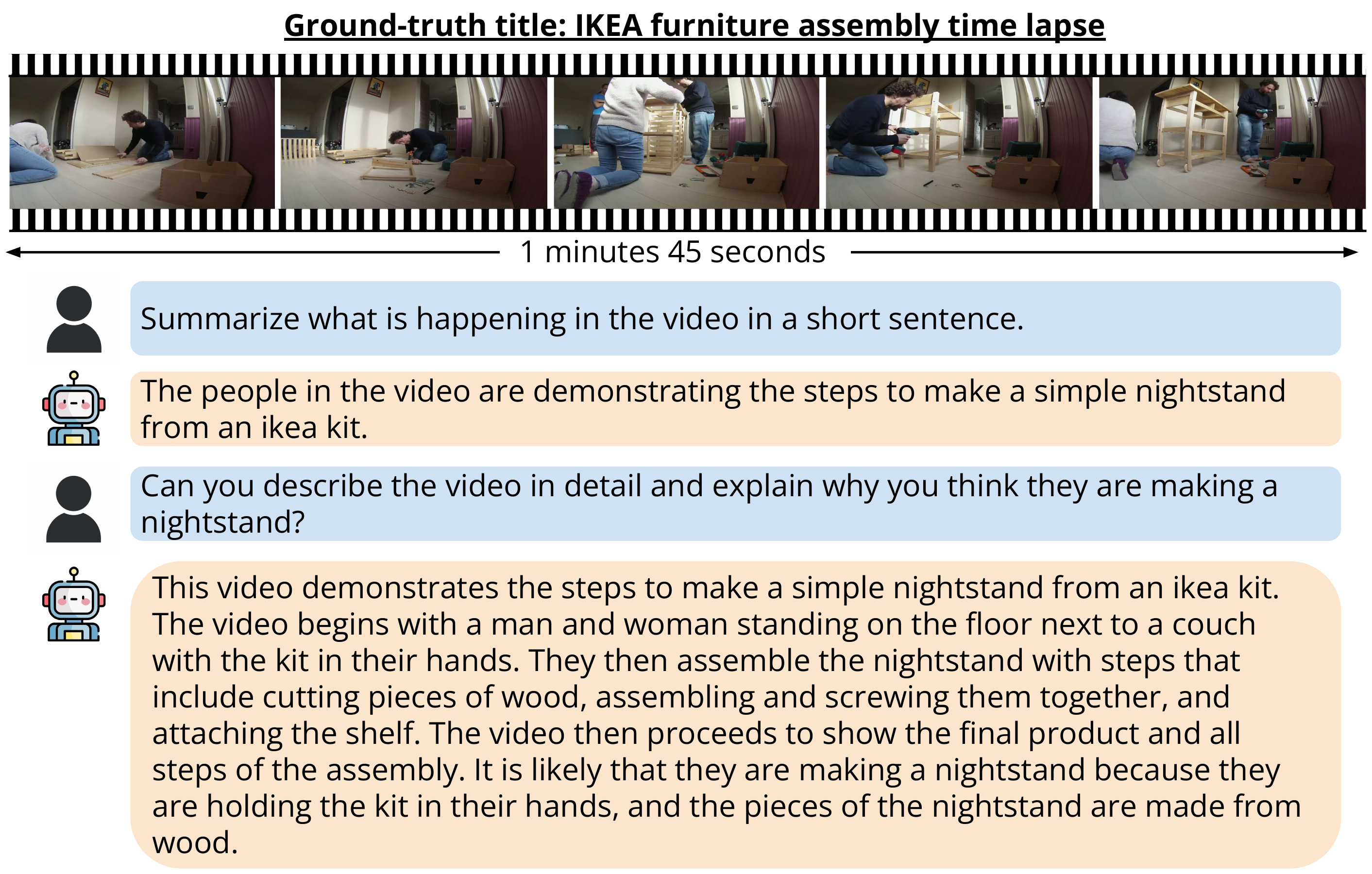}} \\
  \subfloat[b][Video link: \url{https://www.youtube.com/watch?v=62hBgU8XzKo}]
{\includegraphics[width=0.95\linewidth]{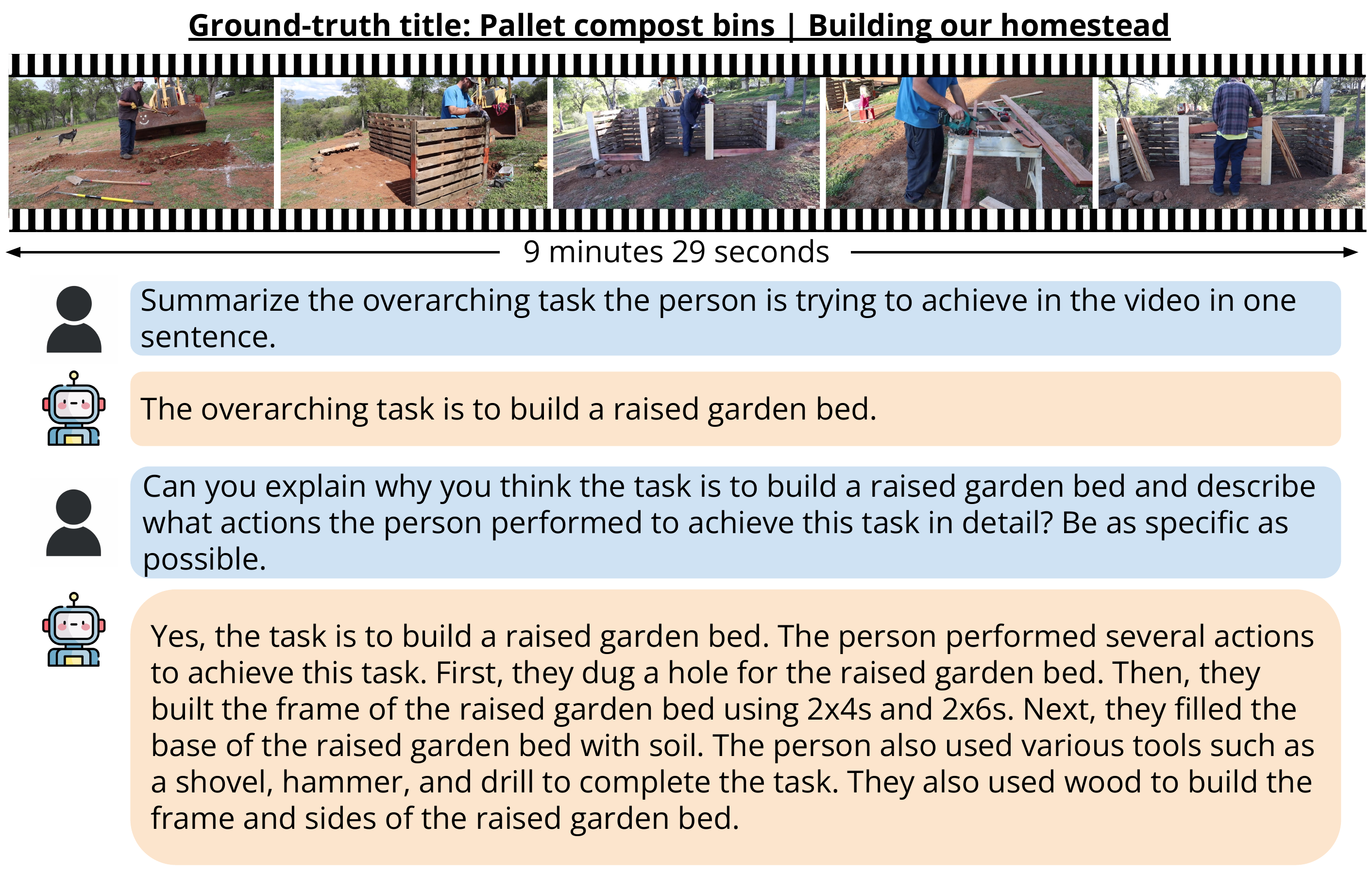}}
  \caption{Sample generations.} \label{fig:sample_conversations_1}
\end{figure*}

\begin{figure*}[t!]
  \centering
  \subfloat[a][Video link: \url{https://www.youtube.com/watch?v=T33BkvAkctY}]{\includegraphics[width=0.95\linewidth]{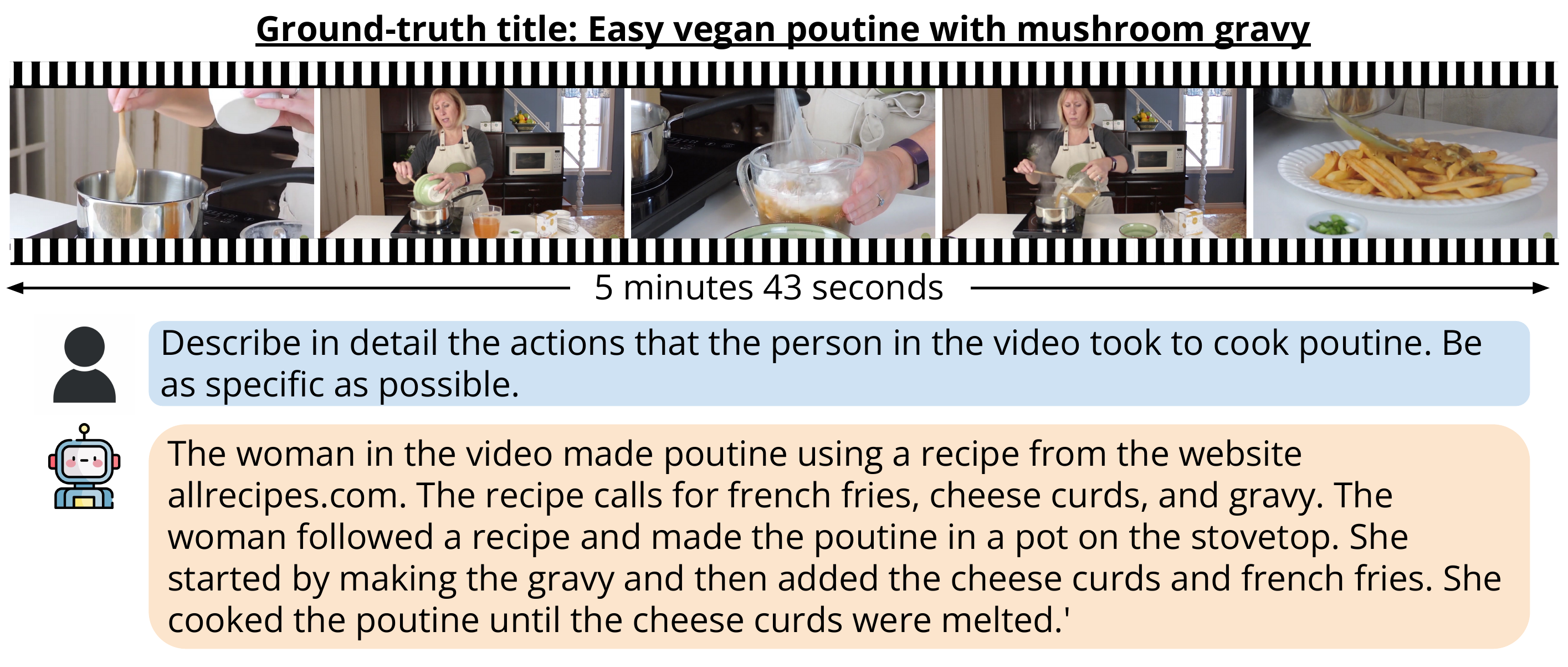} } \\

  \subfloat[b][Video link: \url{https://www.youtube.com/watch?v=0jRg9DRDnrU}]{\includegraphics[width=0.95\linewidth]{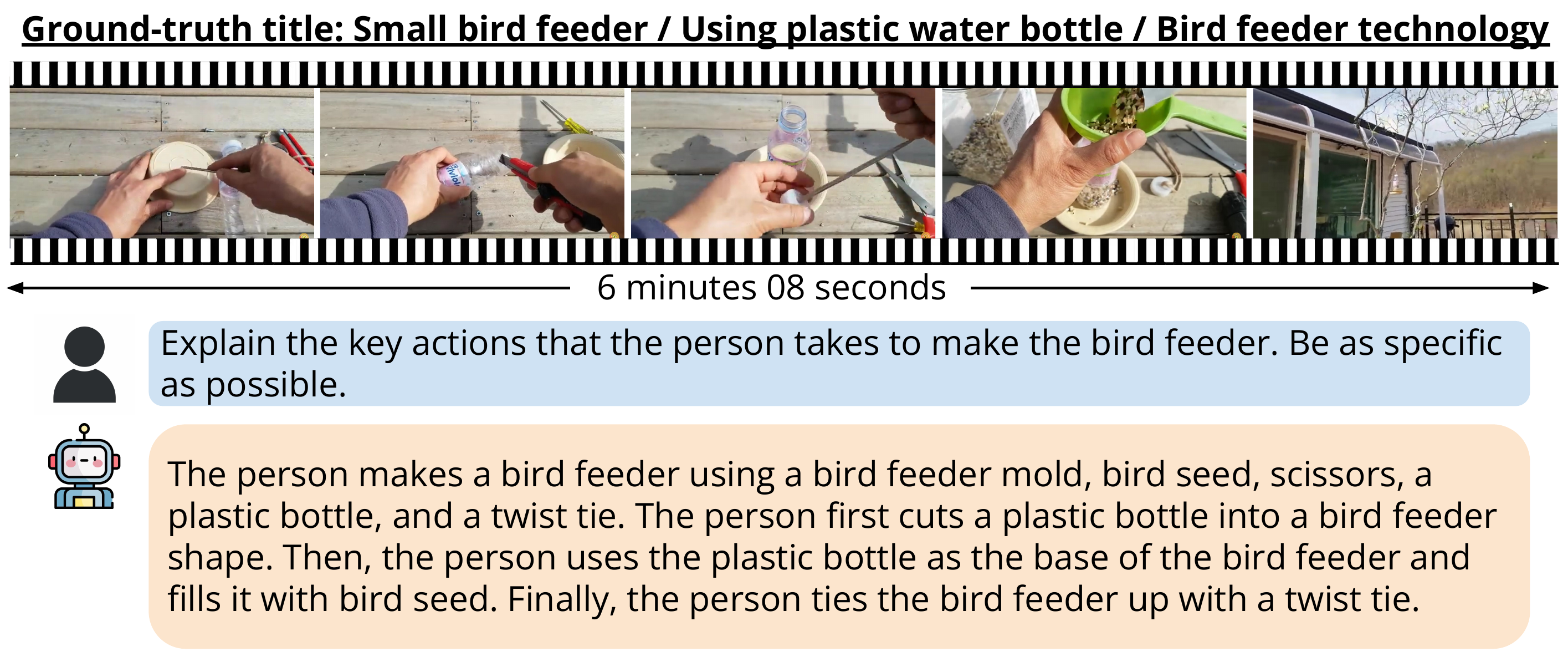} }
  \caption{Sample generations.} \label{fig:sample_conversations_2}
\end{figure*}

\noindent\textbf{Sample conversational generations.} Using our final pretrained \modelabb model, we also provide qualitative visualizations of sample conversations with videos that are randomly downloaded from YouTube. In Figure~\ref{fig:sample_conversations_1}, we observe that our \modelabb model is capable of reasoning about the contextual relationships between multiple short actions to infer reasonable summaries of long videos. For instance, we see that \modelabb is also able to explain the reasoning behind its predictions of making a nightstand and constructing a raised garden bed in Figure~\ref{fig:sample_conversations_1}\textcolor{red}{a} and \ref{fig:sample_conversations_1}\textcolor{red}{b}, respectively. Additionally, we also provide examples of questioning our \modelabb vLLM about important details in long videos in Figure~\ref{fig:sample_conversations_2}. We see that our vLLM is generally able to structure its responses using the correct temporal ordering of the observed actions.


\begin{figure*}[!t] 
\captionsetup[subfigure]{labelformat=empty}
  \centering
  \subfloat[a][(a) Example prediction 1]{\includegraphics[width=\linewidth]{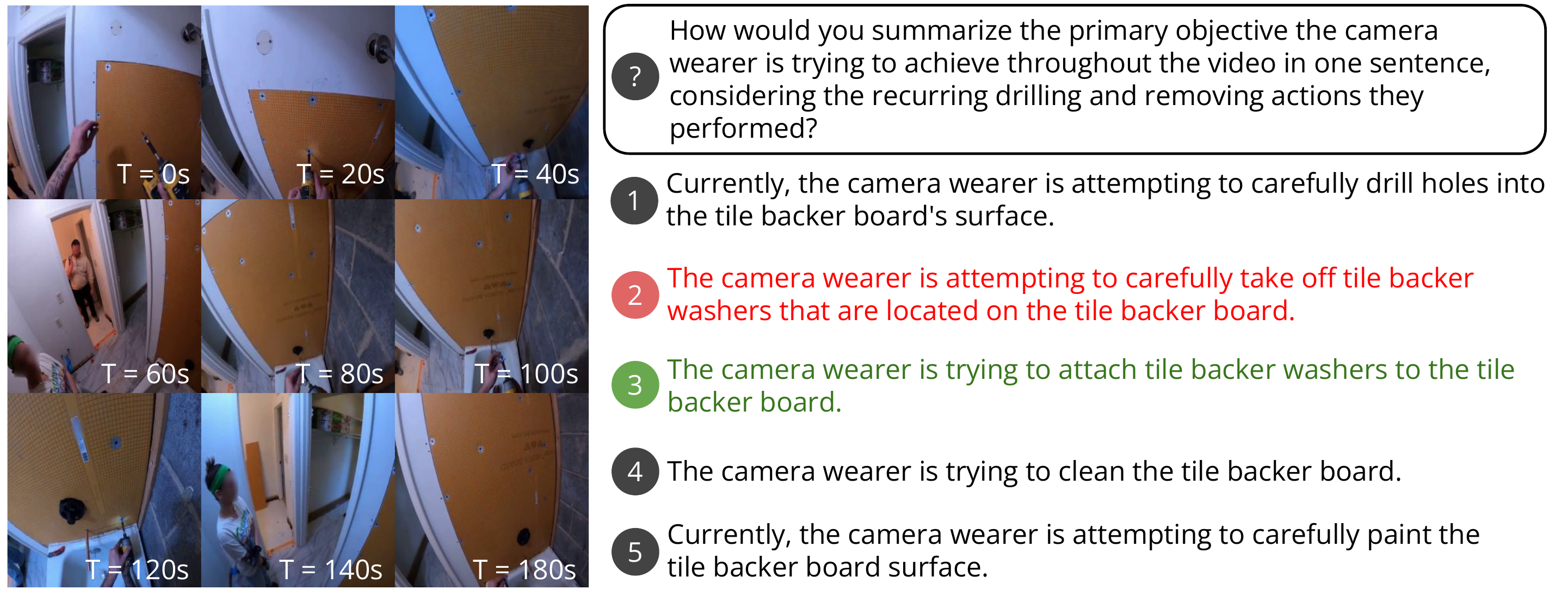}} \\
  
  \subfloat[b][(b) Example prediction 2]{\includegraphics[width=\linewidth]{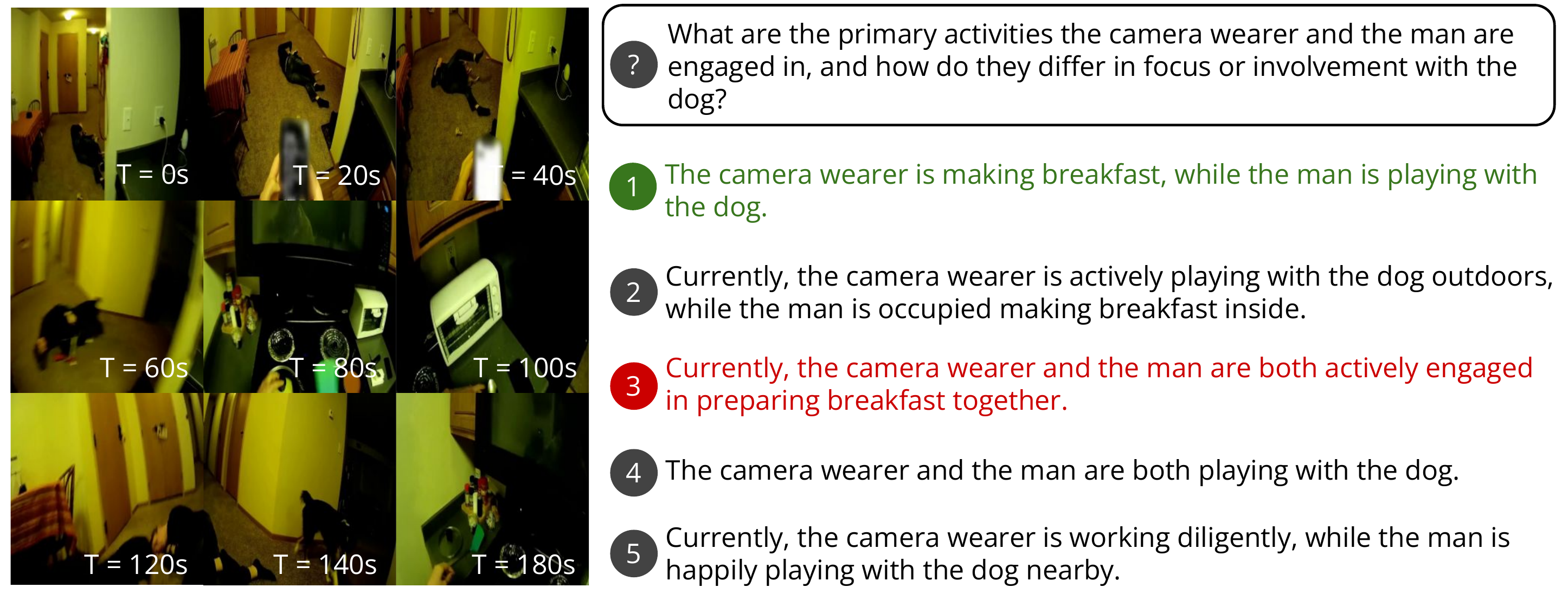}}
  \vspace{30pt}
\end{figure*}

